\documentclass[sigconf]{acmart}

\AtBeginDocument{%
  \providecommand\BibTeX{{%
    \normalfont B\kern-0.5em{\scshape i\kern-0.25em b}\kern-0.8em\TeX}}}

\copyrightyear{2024} 
\acmYear{2024} 
\setcopyright{rightsretained} 
\acmConference[KDD '24]{Proceedings of the 30th ACM SIGKDD Conference on Knowledge Discovery and Data Mining}{August 25--29, 2024}{Barcelona, Spain}
\acmBooktitle{Proceedings of the 30th ACM SIGKDD Conference on Knowledge Discovery and Data Mining (KDD '24), August 25--29, 2024, Barcelona, Spain}\acmDOI{10.1145/3637528.3671848}
\acmISBN{979-8-4007-0490-1/24/08}

\usepackage{amsfonts}
\usepackage{mathtools}
\newtheorem{definition}{Definition}

\usepackage{multirow}
\usepackage{subcaption}
\usepackage{diagbox}
\usepackage{algorithm}
\usepackage{algpseudocode}
\renewcommand{\algorithmicrequire}{ \textbf{Input:}}    
\renewcommand{\algorithmicensure}{ \textbf{Output:}}

\begin{document}

\title{Toward Structure Fairness in Dynamic Graph Embedding: A Trend-aware Dual Debiasing Approach}

\author{Yicong Li}
\authornote{Having equal contribution with the first author.}
\affiliation{
 \institution{The Education University of Hong Kong}
 \country{Hong Kong SAR, China}
 }
\affiliation{
 \institution{University of Technology Sydney}
 \country{Australia}
 }
\email{lyicong@eduhk.hk}
\email{yicong.li@student.uts.edu.au}

\author{Yu Yang}\authornotemark[1]
\authornote{Corresponsing authors.}
\affiliation{
 \institution{The Hong Kong Polytechnic University}
 \country{Hong Kong SAR, China}
 }
\email{cs-yu.yang@polyu.edu.hk}

\author{Jiannong Cao}
\affiliation{
 \institution{The Hong Kong Polytechnic University}
 \country{Hong Kong SAR, China}
 }
\email{jiannong.cao@polyu.edu.hk}

\author{Shuaiqi Liu}
\affiliation{
 \institution{The Hong Kong Polytechnic University}
 \country{Hong Kong SAR, China}
 }
\email{shuaiqi.liu@connect.polyu.hk}

\author{Haoran Tang}
\affiliation{
 \institution{The Hong Kong Polytechnic University}
 \country{Hong Kong SAR, China}
 }
\email{haoran.tang@connect.polyu.hk}

\author{Guandong Xu}\authornotemark[2]
\affiliation{
 \institution{The Education University of Hong Kong}
 \country{Hong Kong SAR, China}
 }
\affiliation{
 \institution{University of Technology Sydney}
 \country{Australia}
 }
\email{gdxu@eduhk.hk}
\email{guandong.xu@uts.edu.au}

\begin{abstract}
Recent studies successfully learned static graph embeddings that are structurally fair by preventing the effectiveness disparity of high- and low-degree vertex groups in downstream graph mining tasks. However, achieving structure fairness in dynamic graph embedding remains an open problem. Neglecting degree changes in dynamic graphs will significantly impair embedding effectiveness without notably improving structure fairness. This is because the embedding performance of high-degree and low-to-high-degree vertices will significantly drop close to the generally poorer embedding performance of most slightly changed vertices in the long-tail part of the power-law distribution. We first identify biased structural evolutions in a dynamic graph based on the evolving trend of vertex degree and then propose FairDGE, the first structurally Fair Dynamic Graph Embedding algorithm. FairDGE learns biased structural evolutions by jointly embedding the connection changes among vertices and the long-short-term evolutionary trend of vertex degrees. Furthermore, a novel dual debiasing approach is devised to encode fair embeddings contrastively, customizing debiasing strategies for different biased structural evolutions. This innovative debiasing strategy breaks the effectiveness bottleneck of embeddings without notable fairness loss. Extensive experiments demonstrate that FairDGE achieves simultaneous improvement in the effectiveness and fairness of embeddings.

\end{abstract}

\ccsdesc[500]{Information systems~Data mining}

\keywords{Dynamic graph embedding; Structural fairness; Degree fairness; Debiased learning; Structural evolution}

\maketitle

\section{Introduction}
\label{sec:intro}
Graph embedding has achieved great success in many applications, such as recommendation \cite{li2024attention}, social inference \cite{sun2023self}, risk assessment \cite{sun2023counter}, etc. Many of these applications are human-centered, where fairness matters a lot \cite{raghavan2021societal}. For example, recommending jobs unbiasedly, making every demographic have the same opportunities, ensuring algorithmic credit scoring for lending does not unintentionally favor certain ethnic or age groups, providing equitable access and connectivity in transportation networks, etc. This drives the research on fair graph embeddings.

Fair graph embedding aims to learn low-dimension vertex/edge representations that will not make disparate treatment or impacts to vertices/edges in downstream graph mining tasks~\cite{kleinberg2018algorithmic}. Disparate treatment, which is direct discrimination, intentionally treats vertices/edges differently based on sensitive attributes \cite{barocas2016big}. Disparate impact is indirect discrimination, where algorithms perform much worse on vertices/edges with sensitive attributes and/or connectivity than on others \cite{barocas2016big}.

Existing works successfully learn fair embeddings for static graphs. Some studies employed adversarial training \cite{dai2021say}, fairness regularizations \cite{fu2020fairness, ge2022explainable, song2022guide}, resampling \cite{tang2023tier, zhao2021hierarchical}, or graph argumentation \cite{yu2022graph} to make sensitive vertex attributes transparent to the embeddings, thus eliminating disparate treatment or impacts to sensitive attributes. Other studies \cite{kose2021fairness, wang2022uncovering, liu2023generalized, tang2020investigating} learned static graph embeddings that are structurally fair by preventing the effectiveness disparity of high- and low-degree vertex groups (i.e., disparate impact) in downstream graph mining tasks, given that vertex degrees of real-world graphs usually follow a long-tailed power-law distribution \cite{barabasi1999emergence}.

Despite the success of learning fair embeddings for static graphs, achieving structure fairness in dynamic graph embedding remains an open problem. Naively re-training the above structurally fair static graph embedding methods from scratch at each timestamp fails to deal with the evolving bias caused by structural changes in dynamic graphs \cite{dai2022towards, li2023early}. As the graph structure evolves, a minority of lower-degree vertices ({\it{Tail}}) may become higher-degree vertices ({\it{Head}}) by actively forming edges with other vertices, while most tail vertices may change slightly. If the model prioritizes fairness by reducing head and tail vertices' effectiveness disparity without considering their evolutionary patterns, the performance of head and tail-to-head vertices will significantly drop close to generally poorer embedding performance of most slightly altered vertices in the long-tail part of the power-law distribution. 
Such performance degradation is empirically demonstrated in Section~\ref{app:sec:dgnn} of the Appendix.
As a result, the effectiveness of learned structurally fair embeddings in downstream tasks is questionable.

In this paper, we study a critical but overlooked problem of learning structurally fair dynamic graph embeddings that are highly effective in downstream graph mining tasks. To achieve this goal, we are confronted with two major challenges. (1) Learning biased structural evolution over time. The structural evolution of head and tail-to-head vertices leaves more linkage-changing information than fluctuation tail vertices, giving rise to the structure bias in the embeddings. Such structure bias changes as the graph evolves, and not all structural evolutions bias embeddings. It is critical to learn biased structural evolution for later debiasing. (2) Debiasing properly to achieve fairness without sacrificing or even improving embedding effectiveness. Debiasing unbiased structural evolution impairs embedding effectiveness without notably improving fairness. Structurally fair embedding algorithms should customize debiasing strategies for different biased structural evolutions such that the effectiveness of embedding is not greatly sacrificed.

In light of these challenges, we first identify three biased structural evolutions in a dynamic graph based on the evolutionary trend of vertex degree, i.e., \textit{Fluctuation-at-Tail (FaT)}, \textit{Tail-to-Head (T2H)}, and \textit{Starting-from-Head (SfH)}, then innovatively propose \textbf{FairDGE}, a structurally \textbf{Fair} \textbf{D}ynamic \textbf{G}raph \textbf{E}mbedding algorithm. To learn biased structural evolution over time, FairDGE jointly embeds the connection changes among vertices and the long-short-term evolutionary trend of vertex degrees by an intermedia training task that classifies the identified biased structural evolutions. Next, we devise a dual debiasing approach for FairDGE to encode fair embeddings that are highly effective in downstream tasks. It first debiases the embeddings via contrastive learning, bringing closer the embeddings with identical biased structural evolutions and penalizing those with different ones. Thanks to the theoretical proof in \cite{wang2022uncovering}, the embedding performance of the FaT vertices will be much improved, resulting in a smaller performance gap with T2H and SfH vertex groups. Meanwhile, we minimize the effectiveness disparity of embeddings of T2H and SfH vertex groups, which are in the head of the power-law distribution and with high performance, to boost fairness further. Consequently, FairDGE breaks the effectiveness bottleneck of embeddings with almost no fairness loss.

Our contributions are highlighted as follows:
\begin{itemize}
    \item \textbf{A new problem.} To the best of our knowledge, we are the first to study the structure fairness problem in dynamic graph embedding and identify three biased structural evolutions that are \textit{Fluctuation-at-Tail (FaT)}, \textit{Tail-to-Head (T2H)}, and \textit{Starting-from-Head (SfH)}.
    \item \textbf{A new and effective approach.} We propose FairDGE, a novel dynamic graph embedding algorithm, that learns biased structural evolutions and then uses a newly devised dual debiasing method for encoding structurally fair embeddings that are highly effective in downstream graph mining tasks.
    \item \textbf{Extensive experiments.} The experimental results demonstrate that FairDGE achieves simultaneous improvement in the effectiveness and fairness of embeddings.
\end{itemize}
\section{Problem Formulation}
\label{sec:problem_def}
In this section, we give the definition of dynamic graphs and formulate the problem of structurally fair dynamic graph embedding.

Firstly, we define a dynamic graph as a snapshot graph sequence.
\begin{definition}
    \label{Def:Dynamic_graph}
    \textbf{A Dynamic Graph.}
    A dynamic graph $\mathcal{G}=(\mathcal{V}, \mathcal{E})$ $=(\mathcal{G}_1, \mathcal{G}_2, ..., \mathcal{G}_T)$ is a sequence of directed or undirected snapshot graphs $\mathcal{G}_t$, where $\mathcal{G}_t=(\mathcal{V}_t,\mathcal{E}_t)$ is a static graph at time $t\in \{1,2,...,T\}$.
    $\mathcal{V}_t$ is a subset of the vertex set $\mathcal{V}=\{v_1,v_2,...,v_{|\mathcal{V}|}\}$.
    An edge $e_{i,j}^t=(v_i^t,v_j^t)\in \mathcal{E}_t$ represents the connection between vertices $v_i^t$ and $v_j^t$ in $\mathcal{G}_t$, where $v_i^t,v_j^t\in \mathcal{V}_t$ and $\mathcal{E}_t$ is a subset of the edge set $\mathcal{E}$.
\end{definition}

Similar to the existing work~\cite{wang2022uncovering, liu2023generalized, dai2021say}, we adopt statistical parity proposed in~\cite{dwork2012fairness} as the fairness metric, as in Definition~\ref{Def:fairness_metric}. In other words, the smaller the intergroup performance disparity, the better the fairness.
\begin{definition}
    \label{Def:fairness_metric}
    \textbf{Fairness Metric.} Given $q$ mutually exclusive vertex groups $\{G_1, ..., G_q\}$ and $G_1\bigcup ...\bigcup G_q=\mathcal{V}$, fairness is achieved when $\mathcal{P}(h_v|v\in G_1) = ... = \mathcal{P}(h_v|v\in G_q)$, where $\mathcal{P}(\cdot)$ indicates the performance of $v$'s embedding $h_v$ in a downstream task.
\end{definition}

Dynamic graph embedding aims to learn low-dimensional representations (i.e., embeddings) for every vertex that preserves its dynamic connectivity changes. The learned embedding should be highly effective in downstream graph mining tasks such as high classification accuracy, small regression errors, high hit rate in recommendation, etc. Achieving structure fairness in dynamic graph embeddings requires the embeddings learned from different structural evolution groups to perform similarly in downstream graph mining tasks.
To this end, we give the formal problem formulation of structurally fair dynamic graph embedding in Definition~\ref{Def:fair_Dynamic_graph_embedding}.

\begin{definition}
    \label{Def:fair_Dynamic_graph_embedding}
    \textbf{Structurally Fair Dynamic Graph Embedding.} Given a dynamic graph $\mathcal{G}=(\mathcal{V}, \mathcal{E})=(\mathcal{G}_1, \mathcal{G}_2, ..., \mathcal{G}_T)$, the vertex $\mathcal{V}$ is divided into $q$ mutually exclusive vertex groups $\{G_1, ..., G_q\}$ based on their structural evolution patterns, which is, $G_1\bigcup ...\bigcup G_q=\mathcal{V}$. The objective is to learn a mapping function $f:v\mapsto h_v \in \mathbb{R}^{dim}$ for $\forall v\in \mathcal{V}$ such that the representation $h_v$ preserves the structural evolution of $v$ in terms of dynamic connectivity changes between $v$ and other vertices in $\mathcal{V}$ over time, achieves high performance in downstream graph mining tasks, and satisfies the fairness requirement defined in Definition~\ref{Def:fairness_metric}, where $dim$ is a positive integer indicating the dimension of $h_v$.
\end{definition}

\section{Identifying Biased Structure Evolutions of Dynamic Graphs}
\label{sec:preliminary_study}

In this section, we introduce intuitions for identifying the biased structural evolution of dynamic graphs and explain the fairness implications for dynamic graph embeddings.

The degree of a vertex represents the number of other vertices connected to it, which indicates its neighborhood structures. As the graph structure evolves, the connections among vertices change accordingly, causing vertices’ degrees to vary. Therefore, degree change is capable of approximating the vertex’s neighborhood structure evolution over time. 

Vertex degrees of real-world graphs usually follow a long-tailed power-law distribution~\cite{barabasi1999emergence}. High- and low-degree vertices are in the head and tail parts of the distribution, respectively. When the vertex degree changes as the graph structure evolves, several evolution patterns will be identified based on the evolving trend of degrees, including \textit{Fluctuation-at-Tail (FaT)}, \textit{Tail-to-Head (T2H)}, \textit{Head-to-Tail (H2T)}, and \textit{Fluctuation-at-Head (FaH)}. FaT is defined as the vertex degree being always lower than a given degree threshold $\theta$ in a time period $t$. In other words, the vertex fluctuates at the tail during $t$. FaH is the opposite in which the vertex's degree remains above $\theta$. T2H is defined as the degree of a vertex changes from below $\theta$ to above $\theta$ during $t$. The T2H vertex's degree is initially below $\theta$ but becomes above $\theta$ during $t$.

We conducted a statistical analysis on six months of data in the Amazon Books dataset containing $64,425$ vertices and $217,163$ edges. Two snapshot graphs are constructed using data from the first three and the last three months. The head and tail groups are divided by a degree threshold of $10$ to identify the evolving trend of degrees across snapshot graphs. The results in Table~\ref{tab:sta_degee_trend} show that the above-identified evolving trend of degrees approximately follows the long-tailed power-law distribution as $90.67\%$ vertices fluctuate at the tail with slight connection changes while only $9.33\%$ vertices’ neighborhood structures vary a lot. This makes dynamic graph embedding algorithms exhibit structure unfairness~\cite{wang2022uncovering, tang2020investigating, liu2023generalized}, favoring highly active structural evolutions (i.e., T2H, H2T, and FaH) too much yet discriminating inactive ones (i.e., FaT). Thus, we call them the biased structural evolutions of the dynamic graph.
Learning structurally fair dynamic graph embeddings is essential and has wide real-world applications, as presented in Section~\ref{app:sec:app_fair_dge} of the Appendix.

In the next section, we will present our structurally fair dynamic graph embedding algorithm to learn the above-identified structural evolution patterns for encoding debiased embeddings toward structure fairness.
Since the number of vertices in H2T and FaH is small, we combine them in a single group called \textit{Start-from-Head (SfH)} in the rest of this paper.
The detailed annotation approach of FaT, T2H, and SfH is presented in the Appendix \ref{sec:annotating_algorithm}.
As for complex structural evolution patterns such as first rising and then falling, first falling and then rising, etc., we leave them to future work.

\begin{table}[h]
\caption{Statistics on the evolutionary trend of vertex degrees.} \label{tab:sta_degee_trend}\vspace{-0.5em}
\resizebox{0.47\textwidth}{!}{%
\begin{tabular}{|c|cc|cc|cc|cc|}
\hline
            & \multicolumn{2}{c|}{FaT}     & \multicolumn{2}{c|}{T2H}    & \multicolumn{2}{c|}{H2T}    & \multicolumn{2}{c|}{FaH}    \\ \hline
Snapshots    & $\mathcal{G}_1$        & $\mathcal{G}_2$         & $\mathcal{G}_1$          & $\mathcal{G}_2$       & $\mathcal{G}_1$          & $\mathcal{G}_2$         & $\mathcal{G}_1$          & $\mathcal{G}_2$        \\ \hline
Avg. Degree  & 1.171& 0.936& 0.869& 9.345& 16.362& 3.447& 29.594& 23.237\\ \hline
Std. Degree  & 1.414& 1.000& 1.732& 13.153& 9.849& 2.646& 23.979& 20.396\\ \hline
Vertex Ratio & \multicolumn{2}{c|}{90.67\%} & \multicolumn{2}{c|}{7.02\%} & \multicolumn{2}{c|}{1.45\%} & \multicolumn{2}{c|}{0.86\%} \\ \hline
\end{tabular}}\vspace{-2em}
\end{table}

\section{Structurally Fair Dynamic Graph Embedding Algorithm}
\label{sec:method}

In this section, we present the details of the FairDGE algorithm to learn structurally fair dynamic graph embeddings. The symbol notations are listed in Table~\ref{app:tab:symbol} in the Appendix.

\subsection{Overall Framework of FairDGE}
\begin{figure*}[t]
    \centering
  \includegraphics[width=0.9\linewidth]{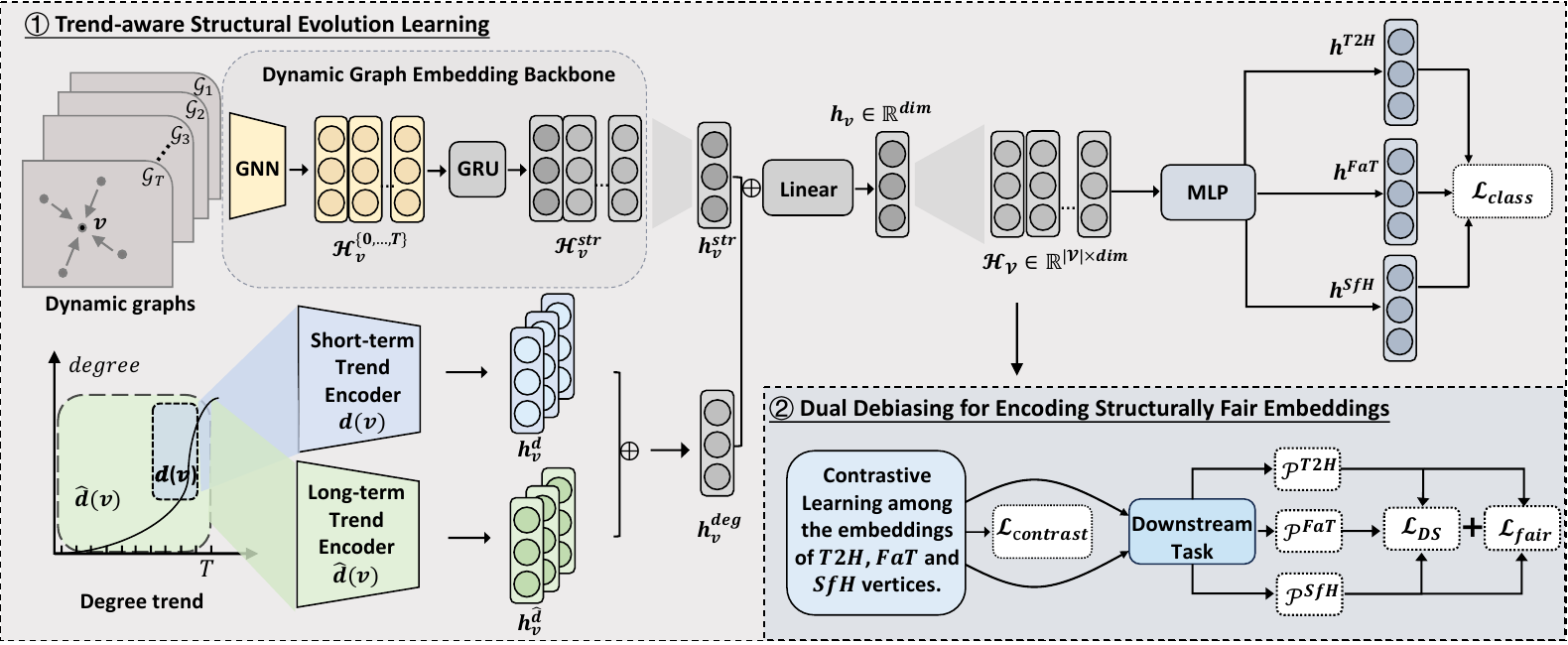}
  \caption{Overall framework of FairDGE.
  }
  \label{fig:method}
\end{figure*}

The overall framework of FairDGE is presented in Figure~\ref{fig:method}.
FairDGE consists of two modules that are trained jointly. We design a trend-aware structural evolution learning module to embed the connection changes among vertices as well as the evolving trend of the corresponding degrees. An intermedia training task that classifies the identified biased structural evolutions (i.e., FaT, T2H, and SfH) is employed to supervise the learning of structural evolutions. 

Another dual debiasing module is devised to encode structurally fair embeddings. It customizes debiasing strategies for FaT, T2H, and SfH vertex groups, respectively. In particular, we first debias the embeddings via contrastive learning, bringing closer the embeddings with identical biased structural evolution and penalizing those with different ones. Hence, the embedding performance of FaT vertices will be significantly improved and close to that of T2H and SfH. Then, we minimize the effectiveness disparity of T2H and SfH vertices' embeddings in downstream tasks to boost fairness further and simultaneously avoid dragging their embedding effectiveness down by FaT vertices with relatively lower performance.

\subsection{Embedding Trend-aware Structural Evolutions in Dynamic Graphs}
We present the details of the trend-aware structural evolution learning module to embed the FaT, T2H, and SfH vertices in a dynamic graph, solving the first challenges identified in Section~\ref{sec:intro}.


\subsubsection{Learning Evolving Trend of Degrees}
\label{sec:degree_modeling}
Calculating the degree difference between the first and last timestamp to measure the trend will lose a lot of information about short-term changes. We first model the degree of vertex $v$ across the snapshot graph sequence $(\mathcal{G}_1, \mathcal{G}_2, ..., \mathcal{G}_T)$ as a time series, denoting as $deg(v)\in \mathbb{R}^{T}$. Then, we devise a long-short-term trend encoder to learn the comprehensive varying patterns of $deg(v)$, including long-term evolving trends and short-term fluctuation patterns.


Specifically, we employ a 3-layer Gate Recurrent Unit (GRU) as the long-term trend encoder, denoting as $\hat{d}(v)$, to learn the long-term evolving patterns of $deg(v)$ as follows:
\begin{equation}
    h_{v}^{\hat{d}} = \hat{d}(v) = GRU(expand(deg(v)))
    \label{eq:deg_trend_encoder}
\end{equation}
where $expand(deg(v))$ expands the dimension of $deg(v)$ from $T\times 1$ to $T \times dim$ by duplication.


To capture and embed the short-term fluctuation patterns of degrees, we employ a 1-d Convolutional Neural Network (CNN) as the short-term trend encoder $d(v)$. 1-d CNN has a good ability to capture the local differences of degrees at several adjacent time points, especially salient degrees \cite{tang2020rethinking}.
\begin{equation}
    h_{v}^{d} = d(v) = CNN(deg(v))
\end{equation}
We set the kernel's size and padding to 3 and 1, respectively. Therefore, $h_{v}^{d}$ keeps the same dimension as $h_{v}^{\hat{d}}$. 

Lastly, we fuse $h_{v}^{d}$ and $h_{v}^{\hat{d}}$ to obtain the trend embedding of $deg(v)$ by a stack-mean operation as shown below.
\begin{equation}
    h_{v}^{deg} = m([h_{v}^{d} \oplus h_{v}^{\hat{d}}])
\end{equation}
where $h_{v}^{deg}$ is the trend embedding of $deg(v)$, $\oplus$ is a stack operator for two embeddings, and $m(\cdot)$ is the mean operation.

\subsubsection{Embedding Structural Evolutions}
Embedding the connection changes among vertices, noting as the structural evolutions, in the snapshot graph sequence $(\mathcal{G}_1, \mathcal{G}_2, ..., \mathcal{G}_T)$ is a typical dynamic graph embedding problem, which has been well studied \cite{yang2023time, yang2021time, pareja2020evolvegcn, you2022roland, sankar2020dysat, zhang2022dynamic}.
Since it is not the main focus of this study and many existing algorithms \cite{yang2023time, yang2021time, pareja2020evolvegcn, you2022roland, zhang2022dynamic, sankar2020dysat, chen2021temporal} are capable of being directly applied, we use a graph neural network (GNN) to embed the snapshot graph at each timestamp and then employ a GRU to learn the sequential changing pattern across the snapshot graph sequence and obtain structural evolution embeddings.
\begin{equation}
    \mathcal{H}_{\mathcal{V}}^{str} = GRU(GNN(\mathcal{G}_1, \mathcal{G}_2, ..., \mathcal{G}_T)))
\end{equation}
where $\mathcal{H}_{\mathcal{V}}^{str}$ is the embedding of all vertices in the dynamic graph. 

We treat this as a dynamic graph embedding backbone that any existing dynamic graph embedding algorithms can be plugged in and learn the structural evolution embeddings.



\subsubsection{Learning Trend-aware Structural Evolutions via An Intermedia Trend Classification Task}
\label{sec:classification_task}
To embed the trend-aware structural evolutions, we fuse the degree-trend embedding of $h_{v}^{deg}$ and the structural evolution embedding $h_{v}^{str}$ for every vertex $v$, as shown in Eq.~(\ref{eq:final_embedding}) where $h_{v}^{str}$ comes from $\mathcal{H}_{\mathcal{V}}^{str}$ for $v\in \mathcal{V}$, $[.;.]$ is a concatenation operation and $g(.)$ is a linear layer.
\begin{equation}
    \label{eq:final_embedding}
    h_{v} = g([h_{v}^{deg};h_{v}^{str}])
\end{equation}

We leverage an intermedia training task to better train $h_{v}$ for retaining the biased structural evolutions FaT, T2H, and SfH.
A two-layer Multi-Layer Perceptron (MLP) is employed to encode the representation, followed by a softmax function to forecast the probability of vertex $v$ belonging to one of the biased structural evolutions, i.e., FaT, T2H, or SfH,
\begin{equation}
    h_{v}^{SfH,T2H,FaT} = MLP(h_{v})
\end{equation}
\begin{equation}
    h_{v,y}=softmax(ReLU(h_{v}^{SfH,T2H,FaT}))
\end{equation}
where $h_{v,y}$ indicates the probability of vertex $v$ that belongs to a class $y \in Y=\{$SfH, T2H, FaT$\}$. 
Lastly, we employ a cross-entropy loss to train $h_{v}$ and $h_{v}^{SfH,T2H,FaT}$, which is
\begin{equation}
    \mathcal{L}_{class} = - \sum_{v\in \mathcal{V}} log \frac{\exp(h_{v,y_v})}{\sum_{y=1}^{Y} \exp(h_{v,y})}
\end{equation}
$h_{v,y_v}$ is a one-hot vector indicating the ground truth biased structural evolutions of $v$, which is manually labeled from the historical data for training. It is fine to set a threshold of head and tail degrees to annotate FaT, T2H, and SfH. However, finding a golden standard to determine the threshold is sometimes difficult due to the ambiguous degree boundary of head and tail vertices. We devise an alternative method, as Algorithm \ref{algorithm:label_strategy} in the Appendix, to annotate FaT, T2H, and SfH based on the slope of degree evolving trends.

\subsection{Dual Debiasing for Encoding Structurally Fair Embeddings}
\label{sec:dual_debiasing}
We devise a dual debiasing module that customizes debiasing strategies for FaT, T2H, and SfH vertices to encode structurally fair embeddings, overcoming the second challenge. The idea is to make the embedding performance of SfH and T2H vertices high and close while improving the embedding performance of FaT vertices as much as possible to narrow the gap with that of SfH and T2H, eventually achieving structure fairness without sacrificing or even improving embedding effectiveness.

R. Wang et al. \cite{wang2022uncovering} had theoretically and empirically proved that contrastive learning will improve the tail vertex's performance and alleviate the performance gap between the head and tail vertex groups, eventually improving fairness.
Inspired by this, we employ contrastive learning to debias the embedding of FaT, T2H, and SfH vertices.
Specifically, for each vertex $v$, we randomly select two vertices $v^{+}$ and $v^{-}$ from $\mathcal{V}$ to form a positive pair $(v,v^{+})$ and a negative one $(v,v^{-})$ in which $v$ and $v^{+}$ are in the same structural evolution group
but $v$ and $v^{-}$ are not.
The objective is to make the embeddings of the positive pairs as close as possible and the representations of the negative pairs as far away as possible, bringing closer the embeddings with identical structural evolutions and penalizing those with different ones.
A contrast loss is proposed in Eq. (\ref{eq:constrast_loss}) to achieve this goal, where $h_v$, $h_{v^{+}}$, $h_{v^{-}}$ are the embeddings of vertex $v$, $v^{+}$ and $v^{-}$, respectively.
\begin{equation}
\label{eq:constrast_loss}
\small
\mathcal{L}_{contrast} = -log \frac{exp(sim(h_{v}, h_{v^{+}}))/ \tau}{exp(\frac{sim(h_{v}, h_{v^{+}})}{\tau}) + \sum_{v^- \in \mathcal{V}^-} exp(\frac{sim(h_v, h_{v^{-}})}{\tau})}
\end{equation}
$sim(h_{v}, h_{v^{+}})$ is measuring the cosine similarity between $h_{v}$ and $h_{v^{+}}$. $\tau$ is a scaling parameter. $\mathcal{V}^-$ is a set of negative vertices with different structural evolutions from $v$.
Thanks to contrastive learning, the embedding performance of FaT vertices will improve, resulting in a smaller performance disparity with T2H and SfH vertices.


Additionally, we alleviate the effectiveness disparity of T2H and SfH vertices' embeddings in downstream tasks for a second debiasing, facilitated by a fairness loss for the embeddings of T2H and SfH vertices in Eq. (\ref{eq:fair_loss}). This loss minimizes the performance disparity of T2H and SfH vertices in downstream tasks, imposing statistical parity as defined in Def.~\ref{Def:fairness_metric}. Eventually, statistical parity equalizes outcomes across T2H and SfH vertices \cite{dwork2012fairness, liu2023generalized}.
\begin{equation}
    \label{eq:fair_loss}
    \mathcal{L}_{fair} = ||\mathcal{L}_{DS}(H_\mathcal{V}^{T2H})-\mathcal{L}_{DS}(H_\mathcal{V}^{SfH})||_2
\end{equation}
$H_\mathcal{V}^{T2H}$ and $H_\mathcal{V}^{SfH}$ are the final learned embeddings of T2H and SfH vertices, respectively. $||\cdot||_2$ denotes the $\ell_2$ norm.
$\mathcal{L}_{DS}(\cdot)$ is a loss function of the downstream tasks, which depends on the actual applications. 
Notablely, FaT vertices are excluded from the second debiasing, avoiding dragging down the performance of T2H and SfH vertices. 

Following \cite{liu2023fair}, we employ the link prediction as the downstream task in this paper. That is
\begin{equation}
    \mathcal{L}_{DS} = -\sum_v^{|\mathcal{V}|} \sum_e^{|\mathcal{E}_v|} x_e \cdot log p_e + (1-x_e)\cdot log(1-p_e)
\end{equation}
$\mathcal{E}_v$ is an edge set of $v$ for training, which contains positive edges $e\in \mathcal{E}_v$ linking vertex $v$ and the same number of negative edges that do not connect to $v$.
$p_e$ is a learned edge existence probability computed by inputting the final embeddings to a Fermi-Dirac decoder \cite{nickel2017poincare}.

Lastly, we jointly minimize all the loss items mentioned above with $\ell_2$ and $\ell_1$ regularization of model parameters $\Theta$, 
\begin{equation}\label{eq:final_loss}
\small
    \mathcal{L} = \gamma_1 \mathcal{L}_{DS} + \gamma_2 \mathcal{L}_{class} + \gamma_3 \mathcal{L}_{contrast} + \gamma_4 \mathcal{L}_{fair} + ||\Theta||_2 + ||\Theta||_1
\end{equation}
where $\gamma_1, \gamma_2, \gamma_3, \gamma_4$ are loss hyperparameters, controlling the proportion of different components in the loss function.
FairDGE will gradually achieve the best trade-off between fairness and model utility in the downstream tasks when jointly optimizing the loss.

\section{Experiments}
We conduct extensive experiments to validate the embedding effectiveness of FairDGE in terms of performance and fairness.

\subsection{Experiment Settings}
\subsubsection{Datasets}

\begin{table}[]
\caption{Stastics of datasets.} \label{tab:dataset}
\resizebox{0.47\textwidth}{!}{%
\begin{tabular}{|l|l|l|l|l|l|l|}
\hline
Dataset      & \# User  & \# Item   & \# Vertex & \# Edges  & Dens.     &   \# Snapshots \\ \hline
Amazon Books & 4,054    & 17,651    & 21,705    & 72,317    & 0.1\%     &   5    \\ \hline
MovieLens    & 2,342    & 5,579     & 7,921     & 477,419   &  3.65\%   &   15    \\ \hline
GoodReads    & 16,701   & 20,823    &   37,524  & 1,084,781 &  0.3\%    &   15    \\ \hline
\end{tabular}}
\end{table}

The experiments are respectively conducted on small-, medium- and large-scale real-world datasets including Amazon Books, MovieLens and GoodReads. 
In the Amazon Books dataset, the user-book interactions from 09 Aug 2006 to 12 Aug 2007 are selected. For the MovieLens dataset, we use the 10M movie ratings from 07 Jan 2001 to 07 Jan 2003. The GoodReads dataset is from an online book community website, and we select the Children's books from 1 Jan 2013 to 30 Dec 2015. The detailed information of datasets is presented in Table \ref{tab:dataset}.

\subsubsection{Baseline Methods}
To comprehensively benchmark our proposed FairDGE, we carefully select eight baseline methods for performance comparison.
Following are the detailed descriptions of the selected baseline methods.

\underline{Graph learning methods:}
\begin{itemize}
    \item GCN \cite{kipf2016semi}: GCN adopts the graph convolutions to learn the graph embeddings. 
    \item GAT \cite{velivckovic2018graph}: GAT leverages masked self-attentional layers to specify different weights to different vertices in a neighborhood.
\end{itemize}

\underline{Dynamic graph learning methods:} 
\begin{itemize}
    \item MPNN-LSTM \cite{panagopoulos2021transfer}: MPNN-LSTM proposes an LSTM-based message passing mechanism to model the vertices evolution in the dynamic graphs. 
    \item EvolveGCN \cite{pareja2020evolvegcn}: EvolveGCN combines the graph convolutional network model with an RNN to capture temporal dynamics in a graph sequence.
\end{itemize}

\underline{Fair graph learning methods:}
\begin{itemize}
    \item FairGNN \cite{dai2021say}: FairGNN effectively mitigates the impact of sensitive attributes in graph data by incorporating fairness constraints during training.
    \item FairVGNN \cite{wang2022improving}: FairVGNN is an extension of the FairGNN model that includes a variational graph autoencoder to learn latent representations and generate fair and diverse outputs in graph data.
    \item DegFairGNN \cite{liu2023generalized}: DegFairGNN is a fair graph neural network to alleviate the structure bias problem using an equal opportunity fair loss. 
    \item TailGNN \cite{liu2021tail}: TailGNN proposes transferable neighborhood translation aiming at closing the performance disparity between head and tail vertices.
\end{itemize}

\subsubsection{Evalution Metrics of Performance}
\label{sec:hr_ndcg}
We adopt the top $k$ Hit Rate (HR@$k$), top $k$ Precision (PREC@$k$), and top $k$ Normalized Discounted Cumulative Gain (NDCG@$k$) as the performance metrics of the recommendation tasks.
\begin{itemize}
    \item \textbf{HR@$k$} calculates the recall ratio of the prediction list,
    \begin{equation}
    \small
        HR@k = \frac{\#Hits@k}{|pos|}
    \end{equation}
    where $|pos|$ is the number of all positive items. $\#Hits@k$ is the number of top $k$ prediction list hit to the positive items.
        
    \item \textbf{PREC@$k$} measures the precision of the prediction list,
    \begin{equation}
    \small
        PREC@k = \frac{\sum_{u\in U} pred_k(u) \bigcap pos(u)}{\sum_{u\in U} pred_k(u)}
    \end{equation}
    where $U$ is the user set, $pred_k(u)$ is the top $k$ prediction list for user $u$, and $pos(u)$ is $u$'s positive item list.
    
    \item \textbf{NDCG@$k$} calculates the normalized ranking of the recall ratio of the prediction list,
    \begin{equation}
    \small
        NDCG@k = \frac{1}{|pos|} \sum_i^{|pos|} \frac{1}{log_2(p_i^k+1)}
    \end{equation}
    where $p_i^k$ is item $i$'s ranking in the top $k$ prediction list.
    
\end{itemize}

\subsubsection{Evaluation Metrics of Fairness}
\label{sec:fair_eva}
We adopt the commonly used ranking discrepancy metrics to quantitatively evaluate fairness. In particular, we set $k\in K=\{1,5,10,15,20,40,60,80,100\}$ and calculate the average Hit Ratio Difference (rHR) and Normalized Discounted Difference (rND) as fairness metric~\cite{yang2017measuring}.

\begin{itemize}
    \item \textbf{Hit Ratio Difference (rHR)} calculates the difference between the T2H items' ratio in the top $k\in K$ candidates and that in the overall ranking. 
    A similar ratio, leading to less rHR, denotes more fairness. Mathematically,
    \begin{equation}
    \small
        rHR = \frac{1}{Z} \sum_k^K \left| \frac{|T2H_k|}{k} - \frac{|T2H|}{|\mathcal{V}|}\right|
    \end{equation} 
    where $|T2H_k|$ denotes the number of T2H vertices in the top $k$ candidate list and $|\mathcal{V}|$ presents the total number of vertices in the dynamic graph. Given $|\mathcal{V}|$ and $|T2H|$, $Z$ is the highest possible value of rHR for normalization. 

    \item \textbf{Normalized Discounted Difference (rND)} calculates the difference between the T2H items' ranking order in the top $k\in K$ candidates and that in the overall ranking. The smaller the rND, the better the fairness.
    \begin{equation}
    \small
        rND = \frac{1}{Z} \sum_k^K \frac{1}{log_2 k} \left|\frac{|T2H_k|}{k} - \frac{|T2H|}{|\mathcal{V}|}\right|
    \end{equation}
    
\end{itemize}

\subsection{Performance and Fairness Comparison}
\label{sec:baseline}

\begin{table*}[]
\caption{FairDGE compares with 8 baselines on performance and fairness.}\vspace{-1em}
\label{tab:baseline}
\resizebox{\textwidth}{!}{%
\begin{tabular}{|l|ccccc||ccccc||ccccc|}
\hline
\multicolumn{1}{|c|}{Datasets}                                                                                                                                                                      & \multicolumn{5}{c||}{Amazon Books}                                                                                                         & \multicolumn{5}{c||}{MovieLens}                                                      & \multicolumn{5}{c|}{GoodReads}                                      \\ \hline
\multicolumn{1}{|c|}{\multirow{2}{*}{Baselines    }}                                                                                                                                                             & \multicolumn{3}{c|}{Performance ($\uparrow$)}                                              & \multicolumn{2}{c||}{Fairness ($\downarrow$)} & \multicolumn{3}{c|}{Performance ($\uparrow$)}                                              & \multicolumn{2}{c||}{Fairness ($\downarrow$)} & \multicolumn{3}{c|}{Performance ($\uparrow$)}                                              & \multicolumn{2}{c|}{Fairness ($\downarrow$)}  \\ \cline{2-16}
\multicolumn{1}{|c|}{}                                                                                                                                                                              & \multicolumn{1}{c}{HR@20}   & \multicolumn{1}{c}{NDCG@20} & \multicolumn{1}{c|}{PREC@20} & \multicolumn{1}{c}{rHR}        & rND        & \multicolumn{1}{c}{HR@20}   & \multicolumn{1}{c}{NDCG@20} & \multicolumn{1}{c|}{PREC@20} & \multicolumn{1}{c}{rHR}        & rND    & \multicolumn{1}{c}{HR@20}   & \multicolumn{1}{c}{NDCG@20} & \multicolumn{1}{c|}{PREC@20} & \multicolumn{1}{c}{rHR}        & rND     \\ \hline
GCN                                                            & \multicolumn{1}{c}{0.0726$\pm$0.034}  & \multicolumn{1}{c}{0.0285$\pm$0.016}  & \multicolumn{1}{c|}{0.0036$\pm$0.002}  & \multicolumn{1}{c}{0.1557$\pm$0.003}     & 0.2614$\pm$0.014     & \multicolumn{1}{c}{0.0904$\pm$0.047}  & \multicolumn{1}{c}{0.0420$\pm$0.029}  & \multicolumn{1}{c|}{0.0045$\pm$0.002}  & \multicolumn{1}{c}{0.3265$\pm$0.020}     & 0.5433$\pm$0.222  &0.0641$\pm$0.071 &	0.0255$\pm$0.031  	&\multicolumn{1}{c|}{0.0032$\pm$0.003}&	0.3708$\pm$0.004  	&0.6865$\pm$0.027    \\ \hline 
 \quad w/ $\mathcal{L}_{fair}$                      & \multicolumn{1}{c}{0.0799$\pm$0.051}  & \multicolumn{1}{c}{0.0313$\pm$0.027}  & \multicolumn{1}{c|}{0.0040$\pm$0.003}  & \multicolumn{1}{c}{0.1550$\pm$0.003}     & 0.2568$\pm$0.020     & \multicolumn{1}{c}{0.0769$\pm$0.105}  & \multicolumn{1}{c}{0.0347$\pm$0.041}  & \multicolumn{1}{c|}{0.0038$\pm$0.005}  & \multicolumn{1}{c}{0.3207$\pm$0.027}     & 0.5720$\pm$0.157  &0.0767$\pm$0.002  	&0.0309$\pm$0.001  	& \multicolumn{1}{c|}{0.0038$\pm$0.000}&0.3703$\pm$0.001  	&0.6651$\pm$0.021    \\ \hline 
 GAT                                                            & \multicolumn{1}{c}{0.0655$\pm$0.010}  & \multicolumn{1}{c}{0.0261$\pm$0.004}  & \multicolumn{1}{c|}{0.0033$\pm$0.001}  & \multicolumn{1}{c}{0.1559$\pm$0.010}     & 0.2715$\pm$0.047     & \multicolumn{1}{c}{0.1109$\pm$0.011}  & \multicolumn{1}{c}{0.0492$\pm$0.003}  & \multicolumn{1}{c|}{0.0055$\pm$0.000}  & \multicolumn{1}{c}{0.3568$\pm$0.002}     & 0.6109$\pm$0.021  &0.0771$\pm$0.015  &0.0296$\pm$0.004  &	\multicolumn{1}{c|}{0.0038$\pm$0.001  }	&0.3839$\pm$0.003  &	0.6403$\pm$0.010    \\ \hline 
 \quad w/ $\mathcal{L}_{fair}$                      & \multicolumn{1}{c}{0.0654$\pm$0.003}  & \multicolumn{1}{c}{0.0253$\pm$0.002}  & \multicolumn{1}{c|}{0.0033$\pm$0.000}  & \multicolumn{1}{c}{0.1552$\pm$0.003}     & 0.2658$\pm$0.009     & \multicolumn{1}{c}{0.0821$\pm$0.050}  & \multicolumn{1}{c}{0.0348$\pm$0.025}  & \multicolumn{1}{c|}{0.0041$\pm$0.002}  & \multicolumn{1}{c}{0.3549$\pm$0.009}     & 0.6038$\pm$0.076   &0.0789$\pm$0.004  	&0.0302$\pm$0.001  &\multicolumn{1}{c|}{0.0039$\pm$0.001  }	&0.3838$\pm$0.001  &	0.6402$\pm$0.013   \\ \hline \hline
 MPNN-LSTM                                                      & \multicolumn{1}{c}{0.1308$\pm$0.037}  & \multicolumn{1}{c}{0.0514$\pm$0.018}  & \multicolumn{1}{c|}{0.0065$\pm$0.002}  & \multicolumn{1}{c}{0.1553$\pm$0.007}     & 0.2615$\pm$0.062     & \multicolumn{1}{c}{0.0359$\pm$0.013}  & \multicolumn{1}{c}{0.0124$\pm$0.005}  & \multicolumn{1}{c|}{0.0018$\pm$0.001}  & \multicolumn{1}{c}{0.3434$\pm$0.025}     & 0.5947$\pm$0.045  &0.0340$\pm$0.056  	&0.0128$\pm$0.022   &\multicolumn{1}{c|}{0.0017$\pm$0.003  }	&0.3683$\pm$0.028  &0.6729$\pm$0.050 \\ \hline
 \quad w/ $\mathcal{L}_{fair}$ & \multicolumn{1}{c}{0.1330$\pm$0.036}  & \multicolumn{1}{c}{0.0520$\pm$0.018}  & \multicolumn{1}{c|}{0.0066$\pm$0.002}  & \multicolumn{1}{c}{0.1553$\pm$0.006}     & 0.2607$\pm$0.039     & \multicolumn{1}{c}{0.0404$\pm$0.022}  & \multicolumn{1}{c}{0.0135$\pm$0.003}  & \multicolumn{1}{c|}{0.0020$\pm$0.001}  & \multicolumn{1}{c}{0.3431$\pm$0.041}     & 0.5932$\pm$0.050   &0.0403$\pm$0.014  	&0.0153$\pm$0.003  	&\multicolumn{1}{c|}{0.0020$\pm$0.001  }&	0.3679$\pm$0.010  &	0.6651$\pm$0.041  \\ \hline 
 EvolveGCN                                                      & \multicolumn{1}{c}{0.0526$\pm$0.024}  & \multicolumn{1}{c}{0.0208$\pm$0.011}  & \multicolumn{1}{c|}{0.0026$\pm$0.002}  & \multicolumn{1}{c}{0.1555$\pm$0.010}     & 0.2630$\pm$0.093     & \multicolumn{1}{c}{0.1707$\pm$0.012}  & \multicolumn{1}{c}{0.0586$\pm$0.017}  & \multicolumn{1}{c|}{0.0085$\pm$0.000}  & \multicolumn{1}{c}{0.3544$\pm$0.050}     & 0.5332$\pm$0.083   &0.0648$\pm$0.096  &	0.0279$\pm$0.001  &	\multicolumn{1}{c|}{0.0033$\pm$0.005  }&	0.3192$\pm$0.038  &	0.5549$\pm$0.031   \\ \hline 
 \quad w/ $\mathcal{L}_{fair}$ & \multicolumn{1}{c}{0.0504$\pm$0.050}  & \multicolumn{1}{c}{0.0199$\pm$0.026}  & \multicolumn{1}{c|}{0.0025$\pm$0.003}  & \multicolumn{1}{c}{0.1552$\pm$0.004}     & 0.2584$\pm$0.008     & \multicolumn{1}{c}{0.1574$\pm$0.088}  & \multicolumn{1}{c}{0.0556$\pm$0.027}  & \multicolumn{1}{c|}{0.0079$\pm$0.004}  & \multicolumn{1}{c}{0.3531$\pm$0.013}     & 0.5261$\pm$0.049  &0.0628$\pm$0.110  	&0.0252$\pm$0.042  &	\multicolumn{1}{c|}{0.0032$\pm$0.006  }&	0.3162$\pm$0.045  &	0.5448$\pm$0.202    \\ \hline \hline
 FairGNN                                    & \multicolumn{1}{c}{0.1792$\pm$0.006}  & \multicolumn{1}{c}{0.0816$\pm$0.047}  & \multicolumn{1}{c|}{0.0090$\pm$0.001}  & \multicolumn{1}{c}{0.1547$\pm$0.006}     & 0.2546$\pm$0.019     & \multicolumn{1}{c}{0.1782$\pm$0.014}  & \multicolumn{1}{c}{0.0713$\pm$0.032}  & \multicolumn{1}{c|}{0.0089$\pm$0.000}  & \multicolumn{1}{c}{0.3787$\pm$0.067}     & 0.6719$\pm$0.028  &0.0695$\pm$0.001  &	0.0242$\pm$0.001  	&\multicolumn{1}{c|}{0.0034$\pm$0.000  }	&0.5149$\pm$0.000  &	0.7096$\pm$0.003    \\ \hline 
 \quad w/ $\mathcal{L}_{fair}$ & \multicolumn{1}{c}{0.1855$\pm$0.071}  & \multicolumn{1}{c}{0.0877$\pm$0.038}  & \multicolumn{1}{c|}{0.0093$\pm$0.004}  & \multicolumn{1}{c}{0.1537$\pm$0.008}     & 0.2515$\pm$0.048     & \multicolumn{1}{c}{0.1809$\pm$0.017}  & \multicolumn{1}{c}{0.0737$\pm$0.017}  & \multicolumn{1}{c|}{0.0091$\pm$0.001}  & \multicolumn{1}{c}{0.3750$\pm$0.051}     & 0.6405$\pm$0.139  &0.0749$\pm$0.005  	&0.0267$\pm$0.004  	&\multicolumn{1}{c|}{0.0037$\pm$0.001  }&	0.5148$\pm$0.074  &	0.7049$\pm$0.008    \\ \hline 
 FairVGNN                                    & \multicolumn{1}{c}{0.1743$\pm$0.127}  & \multicolumn{1}{c}{0.0830$\pm$0.075}  & \multicolumn{1}{c|}{0.0087$\pm$0.006}  & \multicolumn{1}{c}{0.1543$\pm$0.006}     & 0.2516$\pm$0.139     & \multicolumn{1}{c}{0.1385$\pm$0.194}  & \multicolumn{1}{c}{0.0782$\pm$0.322}  & \multicolumn{1}{c|}{0.0069$\pm$0.010}  & \multicolumn{1}{c}{0.3472$\pm$0.038}     & 0.5811$\pm$0.247  &0.0846$\pm$0.033  	&0.0329$\pm$0.006  	&\multicolumn{1}{c|}{0.0043$\pm$0.002  }	&0.4481$\pm$0.120  	&0.6399$\pm$0.411    \\ \hline 
 \quad w/ $\mathcal{L}_{fair}$ & \multicolumn{1}{c}{0.1712$\pm$0.139}  & \multicolumn{1}{c}{0.0816$\pm$0.093}  & \multicolumn{1}{c|}{0.0087$\pm$0.008}  & \multicolumn{1}{c}{0.1539$\pm$0.004}     & 0.2453$\pm$0.067     & \multicolumn{1}{c}{0.1281$\pm$0.088}  & \multicolumn{1}{c}{0.1035$\pm$0.069}  & \multicolumn{1}{c|}{0.0064$\pm$0.005}  & \multicolumn{1}{c}{0.3375$\pm$0.017}     & 0.5683$\pm$0.223   &0.0850$\pm$0.020  &	0.0339$\pm$0.021  	&\multicolumn{1}{c|}{0.0042$\pm$0.001  }	&0.4425$\pm$0.139  	&0.6023$\pm$0.251    \\ \hline 
 DegFairGNN                                    & \multicolumn{1}{c}{0.1692$\pm$0.005}  & \multicolumn{1}{c}{0.0774$\pm$0.018}  & \multicolumn{1}{c|}{0.0085$\pm$0.001}  & \multicolumn{1}{c}{0.1601$\pm$0.012}     & 0.3172$\pm$0.009     & \multicolumn{1}{c}{0.1205$\pm$0.052}  & \multicolumn{1}{c}{0.0481$\pm$0.014}  & \multicolumn{1}{c|}{0.0060$\pm$0.003}  & \multicolumn{1}{c}{0.3612$\pm$0.004}     & 0.6685$\pm$0.105  &0.0526$\pm$0.018  &	0.0201$\pm$0.008  &\multicolumn{1}{c|}{0.0024$\pm$0.003  }&	0.4659$\pm$0.053  &	0.6592$\pm$0.0027    \\ \hline 
 \quad w/ $\mathcal{L}_{fair}$ & \multicolumn{1}{c}{0.1778$\pm$0.029}  & \multicolumn{1}{c}{0.0818$\pm$0.050}  & \multicolumn{1}{c|}{0.0089$\pm$0.002}  & \multicolumn{1}{c}{0.1590$\pm$0.008}     & 0.3103$\pm$0.024     & \multicolumn{1}{c}{0.1208$\pm$0.015}  & \multicolumn{1}{c}{0.0480$\pm$0.012}  & \multicolumn{1}{c|}{0.0060$\pm$0.001}  & \multicolumn{1}{c}{0.3606$\pm$0.016}     & 0.6615$\pm$0.094  &0.0517$\pm$0.019  	&0.0202$\pm$0.006  &	\multicolumn{1}{c|}{0.0026$\pm$0.001  }&	0.4654$\pm$0.019  	&0.6579$\pm$0.022     \\ \hline 
TailGNN                                    & \multicolumn{1}{c}{0.1289$\pm$0.005}  & \multicolumn{1}{c}{0.0631$\pm$0.103}  & \multicolumn{1}{c|}{0.0088$\pm$0.020}  & \multicolumn{1}{c}{0.1543$\pm$0.005}     &   0.2578$\pm$0.028   & \multicolumn{1}{c}{0.1887$\pm$0.075}  & \multicolumn{1}{c}{0.0928$\pm$0.033}  & \multicolumn{1}{c|}{0.0095$\pm$0.004}  & \multicolumn{1}{c}{0.3754$\pm$0.026}     &  0.6564$\pm$0.098  &0.0836$\pm$0.022  	&0.0341$\pm$0.007  	&\multicolumn{1}{c|}{0.0042$\pm$0.001  }	&0.5086$\pm$0.042  	&0.6086$\pm$0.059   \\ \hline 
 \quad w/ $\mathcal{L}_{fair}$ & \multicolumn{1}{c}{0.1316$\pm$0.061}  & \multicolumn{1}{c}{0.0654$\pm$0.086}  & \multicolumn{1}{c|}{0.0082$\pm$0.017}  & \multicolumn{1}{c}{0.1541$\pm$0.005}     &   0.2517$\pm$0.062   & \multicolumn{1}{c}{0.1924$\pm$0.038}  & \multicolumn{1}{c}{0.0946$\pm$0.025}  & \multicolumn{1}{c|}{0.0096$\pm$0.002}  & \multicolumn{1}{c}{0.3692$\pm$0.034}     &    0.6191$\pm$0.163 &0.0773$\pm$0.017  	&0.0319$\pm$0.017  &	\multicolumn{1}{c|}{0.0039$\pm$0.001  }&	0.5082$\pm$0.056  	&0.6072$\pm$0.051   \\ \hline \hline
 FairDGE                                                           & \multicolumn{1}{c}{\textbf{0.2006$\pm$0.033}}  & \multicolumn{1}{c}{\textbf{0.0894$\pm$0.012}}  & \multicolumn{1}{c|}{\textbf{0.0100$\pm$0.002}}  & \multicolumn{1}{c}{\textbf{0.1487$\pm$0.004}}     & \textbf{0.2061$\pm$0.031}     & \multicolumn{1}{c}{\textbf{0.2626$\pm$0.049}}  & \multicolumn{1}{c}{\textbf{0.1330$\pm$0.039}}  & \multicolumn{1}{c|}{\textbf{0.0131$\pm$0.002}}  & \multicolumn{1}{c}{\textbf{0.3189$\pm$0.020}}     & \textbf{0.1824$\pm$0.103}  &\textbf{0.1029$\pm$0.016  }	&\textbf{0.0383$\pm$0.012}	&\multicolumn{1}{c|}{\textbf{0.0052$\pm$0.001  }}	&\textbf{0.3087$\pm$0.033  }&	\textbf{0.3503$\pm$0.006 }   \\ \hline 
 Improvement                                                     & \multicolumn{1}{c}{\textbf{8.18\%}} & \multicolumn{1}{c}{\textbf{1.98\%}}  & \multicolumn{1}{c|}{\textbf{7.91\%}}  & \multicolumn{1}{c}{\textbf{3.27\%}}     & \textbf{15.98\%}    & \multicolumn{1}{c}{\textbf{36.46\%}} & \multicolumn{1}{c}{\textbf{28.50\%}} & \multicolumn{1}{c|}{\textbf{36.33\%}} & \multicolumn{1}{c}{\textbf{0.56\%}}     & \textbf{65.32\%}  &\textbf{21.11\%}	&\textbf{12.30\%}	&\multicolumn{1}{c|}{\textbf{21.09\%}}	&\textbf{2.37\%}	&\textbf{35.69\%}  \\ \hline
\end{tabular}}
\end{table*}

\begin{table*}[t]
\caption{Ablation study of variants on both performance and fairness. $\neg$ means the variant without the following module.}\label{tab:ablation_study}\vspace{-1em}
\resizebox{\textwidth}{!}{%
\begin{tabular}{|l|llllllllll|lll|}
\hline
\multicolumn{1}{|c|}{} & \multicolumn{10}{c|}{Performance($\uparrow$)}                                                                                                                                                                                                                                                                       & \multicolumn{3}{c|}{Fairness($\downarrow$)}                                                         \\ \hline
\multicolumn{1}{|l|}{Variants} & \multicolumn{3}{c|}{HR@\{10,15,20\}}                                                    & \multicolumn{3}{c|}{NDCG@\{10,15,20\}}                                                  & \multicolumn{3}{c|}{PREC@\{10,15,20\}}                                                  & \multicolumn{1}{c|}{Dec.} & \multicolumn{1}{c|}{rHR}    & \multicolumn{1}{c|}{rND}    & \multicolumn{1}{c|}{Inc.} \\ \hline
FairDGE                   & \multicolumn{1}{l|}{\textbf{0.1256}} & \multicolumn{1}{l|}{\textbf{0.1690}}  & \multicolumn{1}{l|}{\textbf{0.2025}} & \multicolumn{1}{l|}{\textbf{0.0704}} & \multicolumn{1}{l|}{\textbf{0.0823}} & \multicolumn{1}{l|}{\textbf{0.0903}} & \multicolumn{1}{l|}{\textbf{0.0126}} & \multicolumn{1}{l|}{\textbf{0.0113}} & \multicolumn{1}{l|}{\textbf{0.0101}} & -                         & \multicolumn{1}{l|}{\textbf{0.1489}} & \multicolumn{1}{l|}{\textbf{0.2075}} & -                         \\ \hline
FairDGE$\neg \mathcal{L}_{fair}$          & \multicolumn{1}{l|}{0.0730} & \multicolumn{1}{l|}{0.1068} & \multicolumn{1}{l|}{0.1396} & \multicolumn{1}{l|}{0.0391} & \multicolumn{1}{l|}{0.0483} & \multicolumn{1}{l|}{0.0562} & \multicolumn{1}{l|}{0.0073} & \multicolumn{1}{l|}{0.0071} & \multicolumn{1}{l|}{0.0070} & \textbf{38.13\%}                   & \multicolumn{1}{l|}{0.1551} & \multicolumn{1}{l|}{0.2699} & \textbf{17.12\%}                   \\ \hline
FairDGE$\neg \mathcal{L}_{class}$                 & \multicolumn{1}{l|}{0.0940} & \multicolumn{1}{l|}{0.1325} & \multicolumn{1}{l|}{0.1672} & \multicolumn{1}{l|}{0.0490} & \multicolumn{1}{l|}{0.0595} & \multicolumn{1}{l|}{0.0679} & \multicolumn{1}{l|}{0.0094} & \multicolumn{1}{l|}{0.0088} & \multicolumn{1}{l|}{0.0084} & \textbf{23.49\%}                   & \multicolumn{1}{l|}{0.1536} & \multicolumn{1}{l|}{0.2605} & \textbf{14.35\%}                   \\ \hline
FairDGE$\neg \mathcal{L}_{contrast}$          & \multicolumn{1}{l|}{0.0646} & \multicolumn{1}{l|}{0.1024} & \multicolumn{1}{l|}{0.1342} & \multicolumn{1}{l|}{0.0334} & \multicolumn{1}{l|}{0.0437} & \multicolumn{1}{l|}{0.0513} & \multicolumn{1}{l|}{0.0065} & \multicolumn{1}{l|}{0.0068} & \multicolumn{1}{l|}{0.0067} & \textbf{42.92\%}                   & \multicolumn{1}{l|}{0.1552} & \multicolumn{1}{l|}{0.2728} & \textbf{17.85\%}                   \\ \hline
FairDGE$\neg$ Deg               & \multicolumn{1}{l|}{0.0681} & \multicolumn{1}{l|}{0.0932} & \multicolumn{1}{l|}{0.1199} & \multicolumn{1}{l|}{0.0361} & \multicolumn{1}{l|}{0.0429} & \multicolumn{1}{l|}{0.0493} & \multicolumn{1}{l|}{0.0068} & \multicolumn{1}{l|}{0.0062} & \multicolumn{1}{l|}{0.0060} & \textbf{45.02\%}                   & \multicolumn{1}{l|}{0.1524} & \multicolumn{1}{l|}{0.2282} & \textbf{6.16\%}                    \\ \hline
FairDGE$\neg$ GRU               & \multicolumn{1}{l|}{0.0456} & \multicolumn{1}{l|}{0.0681} & \multicolumn{1}{l|}{0.0856} & \multicolumn{1}{l|}{0.0254} & \multicolumn{1}{l|}{0.0315} & \multicolumn{1}{l|}{0.0357} & \multicolumn{1}{l|}{0.0046} & \multicolumn{1}{l|}{0.0045} & \multicolumn{1}{l|}{0.0043} & \textbf{60.93\%}                   & \multicolumn{1}{l|}{0.1568} & \multicolumn{1}{l|}{0.2792} & \textbf{19.93\%}                   \\ \hline
\end{tabular}}
\end{table*}

To validate the effectiveness of our proposed FairDGE, we split the Amazon Books, MovieLens, and GoodReads datasets into 5, 15, and 15 snapshots for training and testing, respectively. We regard each user's last interacted item as the test data.
We run our FairDGE and all baselines five times with different random seeds and report the mean and the standard deviation (Std) results in Table \ref{tab:baseline}. Std values are shown magnified by a factor of 10 to save space.

FairDGE performs best on all three datasets, outperforming baseline methods of 1.98\% - 36.46\% in recommendations, and achieves the smallest fairness scores, especially on the rND metric. The smaller the rHR and rND scores, the better the fairness.
GCN and GAT always obtain the worst performance among these baselines because they are simple static graph neural networks and fail to embed the structural evolutions. Although EvolveGCN also gets worse performance on the Amazon Books dataset, it performs well on the MovieLens dataset because the Amazon Books dataset is too sparse for EvolveGCN to learn abundant evolution information via the recurrent architecture. Thanks to generative adversarial network structures, FairGNN and FairVGNN are the best baselines in both recommendation performance and fairness, but they still perform much worse than our FairDGE.

Another observation is that better fairness performance is obtained when equipping the baseline methods with our fairness loss $\mathcal{L}_{fair}$, especially rND. This demonstrates that our fairness loss effectively pushes the \textit{T2H} vertices to get more exposure in the recommendation candidate list, especially to display in the high ranking. At the same time, their recommended performance only dropped slightly, and some even increased. This demonstrates that the dual debiasing approach makes FairDGE successfully overcome the second challenge.

\subsection{Ablation Study of FairDGE}

To study the contributions of each module to performance and fairness, we conducted the ablation study on the Amazon Books dataset as shown in Table \ref{tab:ablation_study}. We compare the \{HR, NDCG, PREC\}@\{10,15,20\} and rHR and rND scores to reveal the performance of recommendation and fairness regarding different variants of FairDGE. The Dec. and Inc. columns denote the comparison with FairDGE about the average decreasing percentage performance scores and the average increasing percentage fairness scores, respectively. The higher the fairness scores, the more discrepancy between the disadvantaged and advantaged vertices, resulting in unfairness.

In Table \ref{tab:ablation_study}, the first row of variants is our proposed FairDGE, achieving the best performance and fairness scores. The second row is FairDGE without the fairness loss $\mathcal{L}_{fair}$. The performance decreases by 38.13\%, and fairness increases by 17.12\%, indicating that both performance and fairness become worse when removing the fairness loss. This validates that our fairness loss not only improves the fairness but also breaks the effectiveness bottleneck and results in better performance. 

The third row in Table \ref{tab:ablation_study} is FairDGE without the intermedia training tasks of classifying the biased structural evolutions, which differentiates the \textit{FaT}, \textit{T2H}, and \textit{SfH} vertex groups. This module influences the most minor performance of the downstream task, but it contributes much to fairness because this classification module aims to separate disadvantaged vertices from all vertices, contributing to the following fairness treatment. 

FairDGE$\neg \mathcal{L}_{contrast}$ denotes a variant without contrastive learning in the dual debiasing. The results show that contrastive learning contributes most in FairDGE to both downstream task performance and fairness because it assists in bringing closer the embeddings with identical biased structural evolution and penalizing those with different ones, further proving the effectiveness of contrastive learning on structural fairness. 

The next variant is removing the degree evolving trend learning module introduced in Section \ref{sec:degree_modeling} from the FairDGE, which contributes to the second-best performance but slightly influences fairness. It is because the supervised classification of biased structural evolutions takes more effort in bias detection, but degree evolving trend learning is more about the dynamic graph embedding backbone for biased structural evolution learning, which limits its influence on performance and fairness. 

The last variant is FairDGE without dynamic structural evolution modeling, which is the GRU layers in the dynamic graph embedding backbone. This variant models the data statically and will not differentiate the biased structural evolutions. Obviously, it performs the worst, demonstrating that dealing with the evolving bias caused by structural changes in dynamic graphs is essential, and our FairDGE solves this problem very well.
In summary, each module of FairDGE is indispensable and contributes to performance and fairness to varying degrees.
\vspace{-2em}
\subsection{Hyperparameter Study}
\label{sec:exp_hyperparameter_study}
\begin{figure}
\centering
\begin{subfigure}{.23\textwidth}
    \centering
    \includegraphics[width=\linewidth]{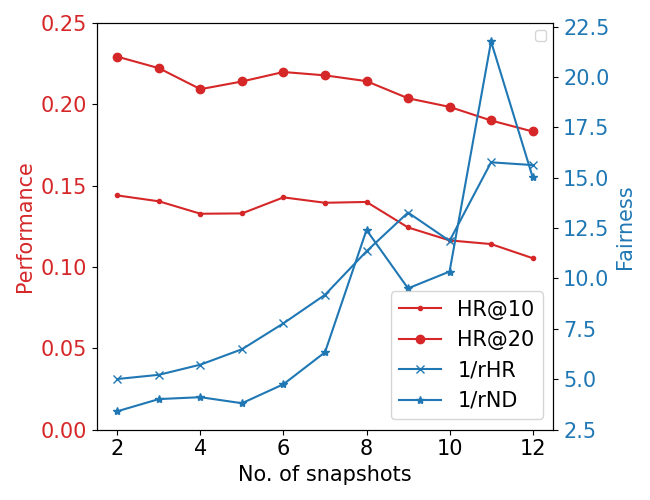}  
    \caption{Snapshots numbers}
    \label{fig:exp_diff_T_Amazon}
\end{subfigure}
~
\begin{subfigure}{.23\textwidth}
    \centering
    \includegraphics[width=\linewidth]{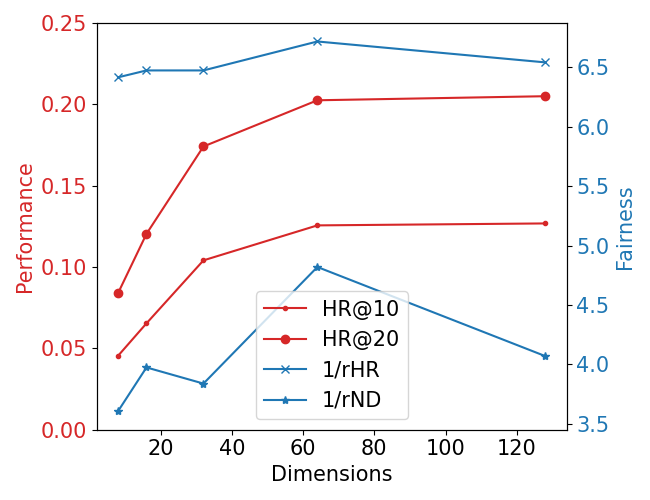}  
    \caption{Model dimensions}
    \label{fig:exp_diff_dim_Amazon}
\end{subfigure}

\begin{subfigure}{.23\textwidth}
    \centering
    \includegraphics[width=\linewidth]{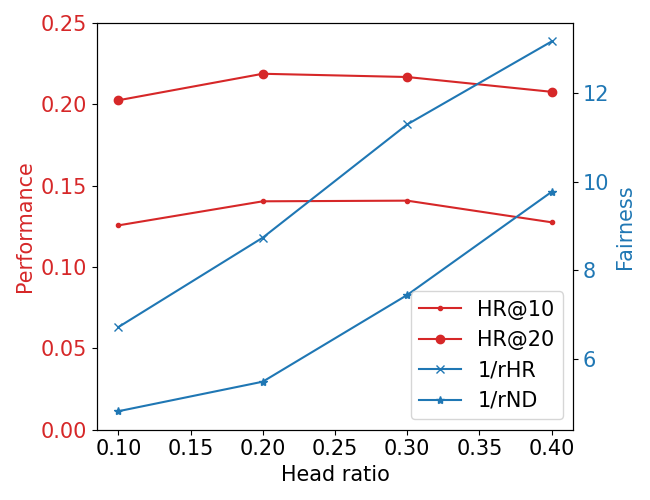}  
    \caption{Head/tail ratios}
    \label{fig:exp_diff_threshold_Amazon}
\end{subfigure}
~
\begin{subfigure}{.23\textwidth}
    \centering
    \includegraphics[width=\linewidth]{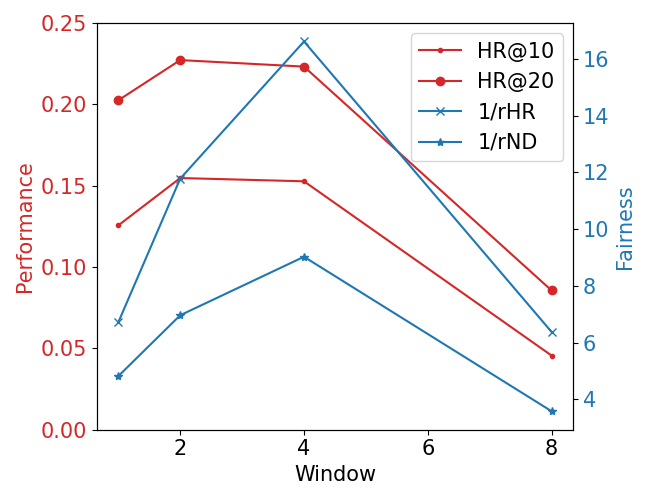}  
    \caption{No. of time-sliding windows}
    \label{fig:exp_diff_window_Amazon}
\end{subfigure}
\caption{Results of hyperparameter testing.}
\label{fig:hyperparams}\vspace{-1.5em}
\end{figure}

To study the influence of different hyperparameters on performance and fairness, we conducted hyperparameter studies on the Amazon Books dataset to test different numbers of snapshots, different model dimensions, different head/tail ratio percentages, and different number of time-sliding windows. For each hyperparameter study, we present the average scores after five runs for a fair comparison. 
The x-axis of Figure \ref{fig:hyperparams} denotes different hyperparameters. The left y-axis displays HR@10 and HR@20 to illustrate the performance trend in recommendation tasks, while the right y-axis exhibits the inversed fairness scores (1/rHR and 1/rND) for convenient visualization. This maintains a consistent interpretation for both sides of metrics, where higher hit rates and higher inversed fairness scores indicate better performance and fairness.

Due to the page limit, we also report the additional hyperparameter testing on loss hyperparameters in Section \ref{app:sec:Hyperparameter_study} of the Appendix.


\subsubsection{Different Number of Snapshots}
To reveal the influence of the number of snapshots on performance and fairness, we report the trend of performance and fairness in Figure \ref{fig:exp_diff_T_Amazon}. The larger the number of snapshots, the shorter the time period within each snapshot. 
We observe a slight decrease in the performance trend as the number of snapshots increases. The reason is that the increased number of snapshots results in sparser snapshot graphs as the number of edges in the snapshot graph decreases. Eventually, the snapshot graph will become sparse, containing rare edge information and deteriorating the performance of each snapshot. 
Additionally, we use GRU in the dynamic graph embedding backbone, which is not good at capturing long-term temporal evolutions. This also causes the performance to drop with increasing snapshots. Fortunately, leveraging more advanced dynamic graph embedding algorithms in the backbone, which widely exist in the community, could solve this problem and further improve the performance of our FairDGE.

From a fairness perspective, more snapshots allow us to obtain finer granular data for capturing the structure evolution in the dynamic graph, helping FairDGE accurately train the degree-evolving trend module to capture change across the evolution snapshots better. Consequently, this benefits our model in effectively identifying different biased vertex groups, eventually improving fairness.
Lastly, the performance and fairness trends reverse as the number of snapshots increases, consistent with the widely accepted notion that performance and fairness tend to be a trade-off.

\subsubsection{Different Dimensions} 
Figure \ref{fig:exp_diff_dim_Amazon} illustrates the variations in performance and fairness as the dimension of embeddings varies in $\{8,16,32,64,128\}$. HR@\{10,20\} exhibits notably low values at the dimension of 8 but experiences a steep increase after 32. Subsequently, the performance reaches its peak at 64. Regarding the fairness scores, it initially performs poorly at the lowest dimension but reaches the best when the dimension of embeddings is 64. However, when the dimension is 128, the fairness scores decrease and become the second-best due to over-training. As a result, we select a default dimension setting of 64 for all the experiments.

\subsubsection{Different head/tail ratio percentages.} 
We vary the head vertex group ratio ranges from 10\% to 40\% and test the impacts on FairDGE's performance and fairness.
Interestingly, we observe that the performance remains consistent across different head/tail ratios in Figure \ref{fig:exp_diff_threshold_Amazon}, highlighting the stability and reliability of our proposed FairDGE algorithm. Although there is a slight peak at a 20\% head vertex split ratio, the overall performance remains unaffected. In contrast, fairness consistently improves with higher head vertex split ratios. By analyzing the proportions of T2H, FaT, and SfH under different head vertex split ratios, we reveal the principle behind this fairness trend. Higher head vertex split ratios result in fewer T2H vertices, reducing the number of vertices used for fairness comparison.

\subsubsection{Different number of time-sliding windows.}
We conduct this experiment to discuss the performance and fairness changing trend if there is a time sliding window to narrow the focusing degree changing range.
Figure \ref{fig:exp_diff_window_Amazon} reveals the performance and fairness changing with the number of time-sliding windows. The time sliding window length is the overall $T$ timestamps. Similarly, if there are $\{1,2,4,8\}$ sliding windows and the number of timestamps is fixed to 8, each time sliding window contains $\{8,4,2,1\}$ snapshots, respectively. The figure shows that the performance and fairness increase as the number of sliding windows grows to 4. As the number of time-sliding windows increases to 8, the model becomes static with 1 snapshot per time-sliding window, resulting in the worst performance and fairness.


\subsection{Convergence and Scalability}

\begin{figure}
\centering
\begin{subfigure}{.24\textwidth}
    \centering
    \includegraphics[width=\linewidth]{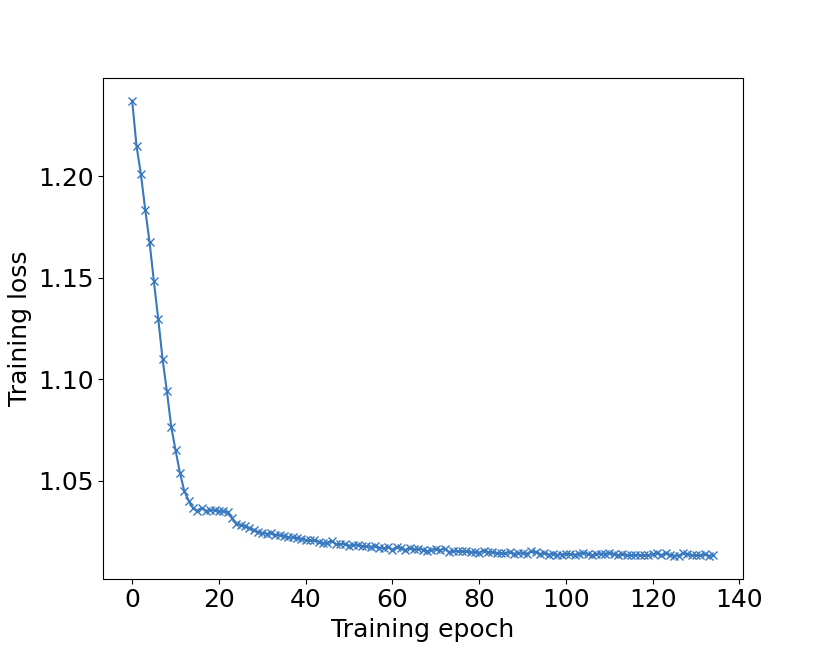}  
    \caption{Training loss w.r.t epoch}
    \label{fig:exp_scalability_training_loss_show}
\end{subfigure}
~\hspace{-0.5em}
\begin{subfigure}{.24\textwidth}
    \centering
    \includegraphics[width=\linewidth]{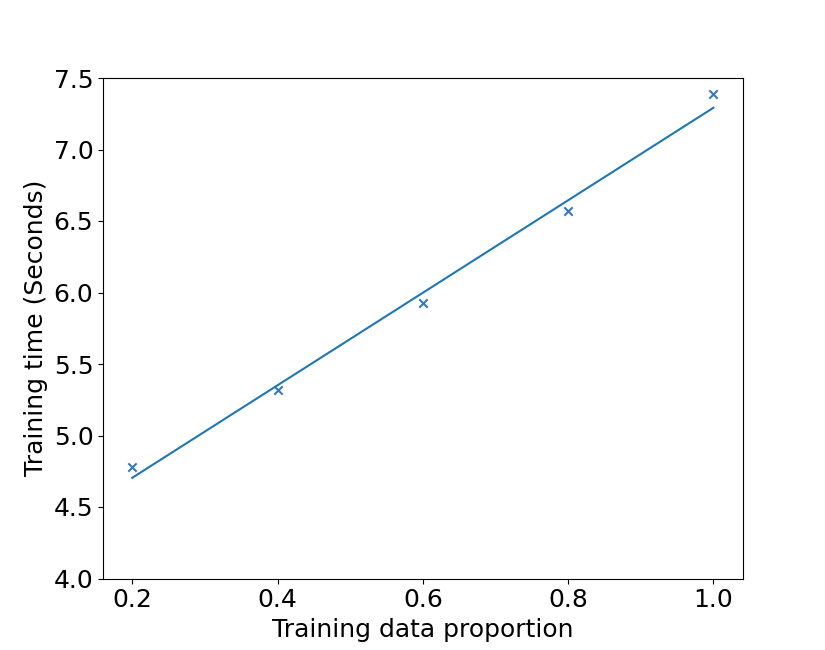}  
    \caption{Training time w.r.t data ratio}
    \label{fig:exp_scalability}
\end{subfigure}\vspace{-1em}
\caption{Convergence and scalability of FairDGE.}\vspace{-1em}
\label{fig:scalability}
\end{figure}

The convergence curve of training FairDGE on the entire Amazon Books dataset is presented in Figure \ref{fig:exp_scalability_training_loss_show}, showing that FairDGE converges at approximately 60 epochs. The x-axis represents the number of training epochs, while the y-axis represents the training loss.
Additionally, we test the scalability of FairDGE by randomly selecting \{20\%, 40\%, 60\%, 80\%, 100\%\} percentages of the training data from the Amazon Books dataset to train FairDGE and report the average training time per epoch in Figure \ref{fig:exp_scalability}. As the volume of training data increases, FairDGE's training time grows almost linearly, demonstrating that FairDGE has good scalability.

\subsection{Effectiveness Analysis of Fairness}
\label{sec:exp_effect_fairness}

\begin{figure}
\centering
\begin{subfigure}{.245\textwidth}
    \centering
    \includegraphics[width=\linewidth]{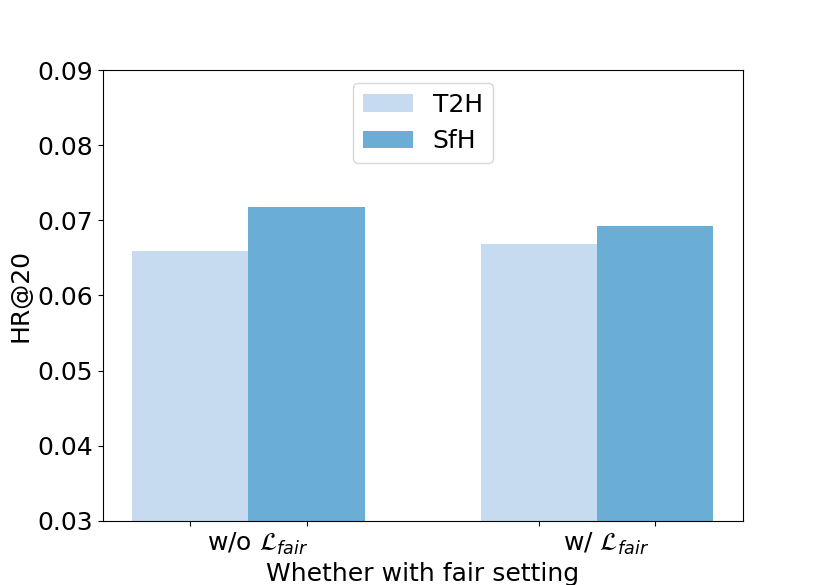}  
    \caption{GCN}
    \label{fig:gcn_truevsnon}
\end{subfigure}
~\hspace{-1.5em}
\begin{subfigure}{.245\textwidth}
    \centering
    \includegraphics[width=\linewidth]{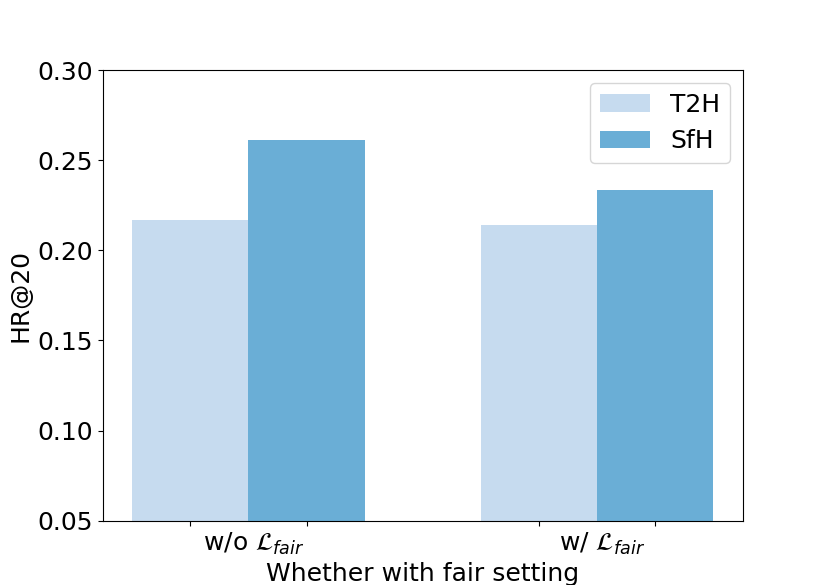}  
    \caption{FairDGE}
    \label{fig:FairDGE_truevsnon}
\end{subfigure}\vspace{-1em}
\caption{Effectiveness of fairness loss $\mathcal{L}_{fair}$. 
}
\label{fig:truevsnon}\vspace{-1.5em}
\end{figure}
In this section, we first discuss the effectiveness of fairness loss $\mathcal{L}_{fair}$ on the T2H and SfH vertex groups and then reveal the superiority of capturing structural evolution for debiasing against the conventional high- and low-degree-based fairness. 

\subsubsection{Effectiveness of Fairness Loss $\mathcal{L}_{fair}$ on The T2H and SfH Vertex Groups.}
To demonstrate the effectiveness of our proposed fairness loss $\mathcal{L}_{fair}$ on the disadvantage group (\textit{T2H}) and advantage group (\textit{SfH}), we analyze the embedding performance in downstream tasks. The closer the performance of T2H and SfH vertex groups, the better the fairness. Specifically, we compare the HR@20 scores of all the biased vertex groups with and without the fairness loss $\mathcal{L}_{fair}$ on GCN and FairDGE. The results are shown in Figure \ref{fig:truevsnon}, where the x-axis denotes whether the fairness loss is used and the y-axis represents the HR@20 scores for performance comparison. 
In each result set, the two bars represent the HR@20 scores of \textit{T2H} and \textit{SfH}, respectively.
We observe that, without $\mathcal{L}_{fair}$, GCN exhibits a significant performance gap between \textit{T2H} and \textit{SfH}, and the performance is much close after equipping $\mathcal{L}_{fair}$. A similar phenomenon is observed in FairDGE. This indicates that the fairness loss $\mathcal{L}_{fair}$ effectively narrows the performance gap between the \textit{T2H} and \textit{SfH} vertex groups, thereby improving fairness.


\begin{figure}
\centering
\begin{subfigure}{.24\textwidth}
    \centering
    \includegraphics[width=\linewidth]{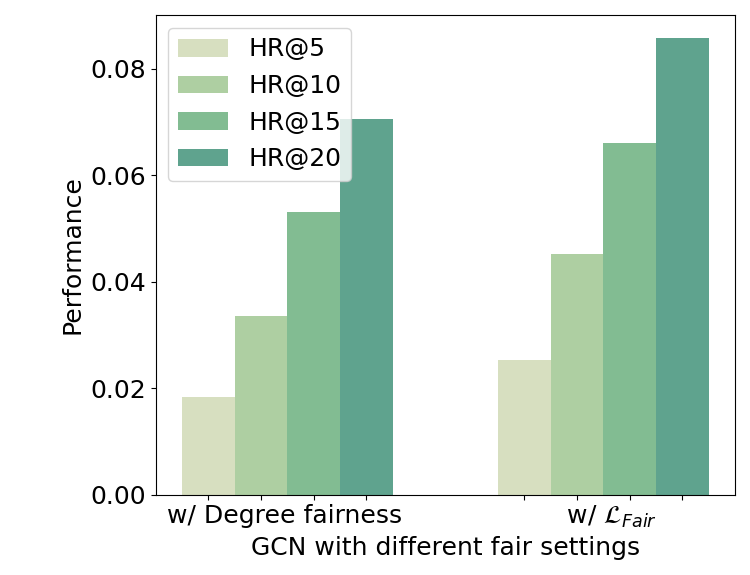}  
    \caption{GCN.}
    \label{fig:gcn_orifair_ourfair}
\end{subfigure}
~\hspace{-0.5em}
\begin{subfigure}{.24\textwidth}
    \centering
    \includegraphics[width=\linewidth]{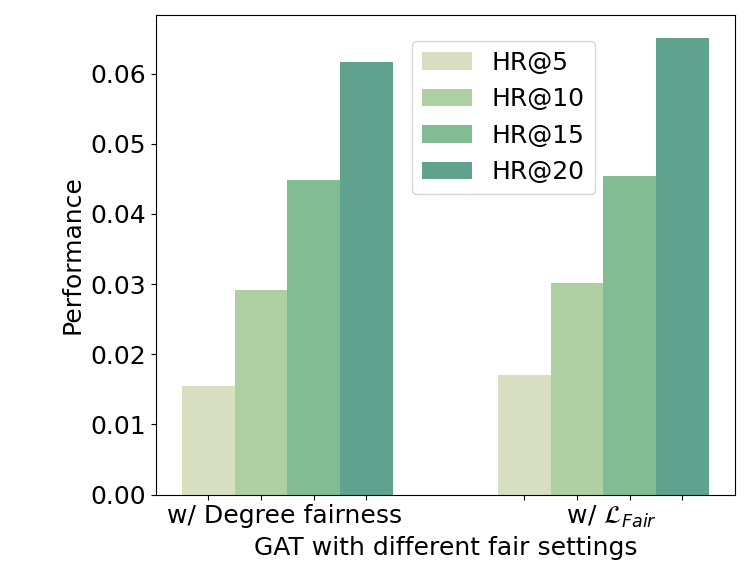}  
    \caption{GAT.}
    \label{fig:gat_orifair_ourfair}
\end{subfigure}\vspace{-1em}
\caption{Effectiveness of debiasing structural evolutions.}
\label{fig:orifair_ourfair_performance}
\end{figure}


\begin{table}[t]
\vspace{-1em}
\caption{Fairness of debiasing structural evolutions.}\vspace{-0.5em}
\label{tab:preliminary_experiment}
\begin{tabular}{|l|l|l|}
\hline
\multicolumn{1}{|c|}{} & rHR ($\downarrow$)   & rND ($\downarrow$)   \\ \hline
GCN w/o Fairness           & 0.1559 & 0.2627 \\ 
\qquad \ w/ Degree Fairness       & 0.1553 \small{(0.4\% $\downarrow$)} & 0.2578  \small{(2\% $\downarrow$)}\\ 
\qquad \ w/ $\mathcal{L}_{fair}$   & \textbf{0.1547} \small{(\textbf{0.8\%} $\downarrow$)}& \textbf{0.2551} \small{(\textbf{3\% $\downarrow$})}\\ \hline \hline 
GAT w/o Fairness           & 0.1569 & 0.2765 \\ 
\qquad \ w/ Degree Fairness       & 0.1557 \small{(0.8\% $\downarrow$)}& 0.2711 \small{(2\% $\downarrow$)}\\ 
\qquad \ w/ $\mathcal{L}_{fair}$   & \textbf{0.1550} \small{(\textbf{1.2\% $\downarrow$})}& \textbf{0.2649} \small{(\textbf{4\% $\downarrow$})}\\ \hline
\end{tabular}\vspace{-1.5em}
\end{table}

\subsubsection{Effectiveness of Capturing Structural Evolution for Debiasing against Degree Fairness.}
Degree fairness prevents the performance disparity of head- and tail-degree vertex groups.
To further validate that capturing structural evolution for debiasing is better than static degree fairness, We divide vertices into head- and tail-degree vertex groups to train GCN and GAT with fairness loss $\mathcal{L}_{fair}$ and compare the performance with GCN and GAT trained on the T2H and SfH vertex groups. The results in Figure~\ref{fig:orifair_ourfair_performance} and Table~\ref{tab:preliminary_experiment} validate that capturing structural evolution to debias dynamic graph embedding is capable of achieving fairness without sacrificing and even improving embedding performance, which degree fairness cannot. These results also demonstrate that our debiasing method can be used as a general fairness plug-in, enabling existing dynamic graph embedding algorithms to achieve structural fairness.


\vspace{-0.5em}
\section{Related Work}

\subsection{Toward Fairness in Static Graph Learning}


Although learning structurally fair dynamic graph embedding remains an under-explored area, several studies have successfully learned fair embeddings for static graphs \cite{wang2024reinforced}, eliminating disparate treatment or impacts to sensitive graph attributes and structures.
To deal with the sensitive graph attributes, various methodologies, such as adversarial learning~\cite{dai2021say, wang2022improving}, resampling techniques~\cite{tang2023tier, zhao2021hierarchical}, and fair-aware loss constraints~\cite{fu2020fairness, ge2022explainable}, have been developed to mitigate bias and promote fairness. 
Specifically, \cite{dai2021say, wang2022improving} adopts adversarial learning to make the discriminators ambiguous about the sensitive feature distinguish, resulting in a fair graph neural network to treat different biased groups equally. To alleviate bias, \cite{rahman2019fairwalk} resamples the vertices to balance the distribution of the sensitive features. Some studies \cite{ge2022explainable, fu2020fairness} propose adding fair constraints to the loss for learning fair graph representations. However, none of them address structure fairness, e.g., degree fairness, in graph learning.

To address the bias introduced by the graph structure, early work revealed that the long-tail distribution of vertex degree caused fairness issues due to the poor embedding performance of tail vertices \cite{liu2021tail, liu2020towards}. Few-shot learning \cite{zhang2020few, wu2022hierarchical}, contrastive learning \cite{wang2022uncovering}, resample \cite{wang2023characterizing, zhao2021hierarchical}, data augmentation \cite{yu2022graph} approaches were proposed to narrow the performance gap between head and tail vertices. TailGNN \cite{liu2021tail} and Meta-Tail2vec \cite{liu2020towards} are the representative works that improve the expressivity of GNN on tail vertices, which surprisingly improves the fairness. Recently, DegFairGNN \cite{liu2023generalized} defined the degree fairness problem in static graph learning for the first time and proposed a fairness constraint to debias the imbalanced degrees of vertices. However, none of the existing studies account for the problem of structure fairness in dynamic graph learning. Our work fills this research gap.



\vspace{-0.5em}
\subsection{Toward Fairness on Dynamic Euclidean Data}
Fairness is a critical problem in many scenarios where time and dynamics matter, including dynamic allocation \cite{si2022enabling}, job prediction \cite{scher2023modelling}, and so on. Therefore, some research works \cite{scher2023modelling, tang2023tier, si2022enabling} alleviate the bias by dynamic solutions, including Markov Decision Processes (MDPs) \cite{zimmer2021learning}, adaptive threshold policy \cite{sinclair2022sequential} and so on. The main research solutions cover fairness accumulation across timestamps \cite{si2022enabling} and long-term fairness increment \cite{tang2023tier}. However, research on fair graph learning or graph embeddings in dynamic scenarios, e.g., dynamic graph embedding \cite{tang2023dynamic}, is scarce. Different from the above-mentioned approaches that isolate the fairness for each timestamp, our study takes into account the evolving trend of vertices' degrees and incorporates graph structural evolution to learn fair representations in dynamic graphs.


\vspace{-0.5em}
\section{Conclusion}

In this paper, we investigate the structure fairness problem in dynamic graph embedding for the first time. We empirically validate that structural evolutions in a dynamic graph approximately follow a long-tailed power-law distribution. This makes dynamic graph embedding algorithms exhibit structure unfairness. We innovatively propose a structurally Fair Dynamic Graph Embedding algorithm, namely FairDGE, which first learns trend-aware structural evolutions and then encodes structurally fair embeddings by a novel dual debiasing approach. 
FairDGE can be used as a general fairness plug-in, enabling existing dynamic graph learning algorithms to generate structurally fair embeddings.


\section*{Acknowledgement}
This work was partially conducted at the Research Institute for Artificial Intelligence of Things (RIAIoT) at PolyU. It is supported by the Hong Kong Research Grants Council (RGC) under the Theme-based Research Scheme with Grant No. T43-513/23-N and T41-603/20-R, and under Research Impact Fund with Grant No. R5034-18. This work is also supported by the Australian Research Council (ARC) under Grant No. DP220103717, and LE220100078.


%

\clearpage


\clearpage

\appendix
\vspace{1em}
\noindent{\Large{\textbf{APPENDIX}}}

\vspace{-1em}
\section{An Application of Fair Dynamic Graph Embedding}
\label{app:sec:app_fair_dge}

Learning fair embeddings for dynamic graphs is essential and has wide real-world applications. Taking user-item recommendations as an example, suppose we merely train a recommendation model without considering the fairness between head (popular) and tail (unpopular) groups of items in the long-tailed power-law distribution. A handful of head (popular) items will easily dominate the models, leading to a much higher probability of being recommended than the tail (unpopular) ones. As a result, the variety of items recommended to users is greatly reduced. Degree fair static graph embedding solves this issue by minimizing the performance disparity of head and tail items, making them have approximately equal probability of being recommended. However, not all tail (unpopular) items deserve the same recommendation probability as the head (popular) ones. Some tail items are of poor quality and users usually dislike them. If the model recommends these low-quality items to users as frequently as it recommends high-quality items, it will seriously impair the quality of service. Therefore, we argue that only a subset of high-quality tail items deserves the same recommendation probability as the head (popular) items. These high-quality tail items will gradually be liked by users and become popular. In other words, they gradually move from tail to head. This phenomenon is also observed in the Amazon Book dataset, as the statistics of tail-to-head (T2H) vertices in Table~\ref{tab:sta_degee_trend} in Section~\ref{sec:preliminary_study}. Making T2H items have a similar recommendation probability as the head ones can increase their exposure and shorten their time to become popular. We abstract this as a structure fairness problem in dynamic graph embedding.


\section{Performance Degradation of Existing Structural Fairness Methods in Dynamic Graph Embedding}
\label{app:sec:dgnn}

Given that the vertex degree of real-world graphs usually follows a long-tailed power-law distribution, existing structural fairness methods prevent the performance disparity of high- and low-degree vertex groups in downstream graph mining tasks without considering the evolving trend of degree. When applying the structural fairness methods in dynamic graph embedding, the tail-to-head vertices will be treated as head ones. Due to the limited connection information carried by the tail vertices, their embedding performance is usually much worse than that of the head (including tail-to-head) vertices and is difficult to boost. When the model minimizes the performance disparity between tail and head vertices to achieve fairness, the performance of head vertices will significantly drop close to the generally poorer embedding performance of the tail vertices. 

Our FairDGE first debiases the embeddings via contrastive learning, increasing the performance of fluctuation-at-tail (FaT) vertices and resulting in a smaller performance gap with tail-to-head (T2H) and starting-from-head (SfH) vertices. In addition, we eliminate the performance disparity of T2H and SfH vertices’ embeddings in downstream tasks for a second debiasing. Notably, FaT vertices are excluded from the second debiasing so that they can pursue the best performance in downstream tasks and avoid dragging down the performance of T2H and SfH vertices. By making the embedding performance of SfH and T2H vertices high and close while improving the embedding performance of FaT vertices as much as possible to narrow the gap with that of SfH and T2H, our approach successfully achieves structure fairness without sacrificing or even improving embedding effectiveness.

\begin{table}[t]
\caption{HR@20 of existing structural fairness against trend-aware fairness in dynamic graph embedding.}\label{app:tab:dgnn_fair}\vspace{-1em}
\begin{tabular}{|l|c|c|}
\hline
\textbf{} & w/ Degree fairness & w/ $\mathcal{L}_{fair}$ \\ \hline
MPNN-LSTM & 0.1285             & 0.1297                  \\ \hline
FairDGE   & 0.2010             & 0.2025                  \\ \hline
\end{tabular}
\end{table}

We conduct an experiment on the Amazon Books dataset to empirically demonstrate the performance degradation of head (including tail-to-head) vertices when applying existing degree fairness methods in dynamic graph embedding. We set a degree threshold of 4 to determine the head and tail vertices and annotate the FaT, T2H, and SfH vertices in dynamic graphs. We first train MPNN-LSTM~\cite{panagopoulos2021transfer}, which is a dynamic graph embedding method, and FairDGE on head- and tail-degree vertex groups and report the HR@20 scores in the column of w/~Degree fairness in Table~\ref{app:tab:dgnn_fair}. Next, we train MPNN-LSTM and FairDGE with our designed fairness loss $\mathcal{L}_{fair}$ in Eq. (\ref{eq:fair_loss}) on FaT, T2H, and SfH groups. The HR@20 scores are shown in the column of w/~$\mathcal{L}_{fair}$ in Table~\ref{app:tab:dgnn_fair}. The HR@20 scores of both MPNN-LSTM and FairDGE with traditional degree fairness are lower than that of using our proposed fairness approach which considers the evolving trend of degrees in the dynamic graph. This validates the performance degradation when applying existing structural fairness methods in dynamic graph embedding.

\section{Symbol Descriptions}
Table \ref{app:tab:symbol} is the symbol description used in this manuscript. \vspace{-1em}

\begin{table}[h]
    \caption{Symbol descriptions.}
    \label{app:tab:symbol}\vspace{-1em}
    \centering
    \resizebox{0.47\textwidth}{!}{%
    \begin{tabular}{c|l}
    \hline
        Symbols & Descriptions \\ \hline
        $\mathcal{G}_t=(\mathcal{V}_t,\mathcal{E}_t)$ & Static snapshot graph, vertex set, edge set at time $t$\\
        $\mathcal{G}=(\mathcal{V}, \mathcal{E})$ & Dynamic graph, vertex set, edge set\\
        $h\in \mathcal{H}$ & A representation vector (embedding)\\
        $\mathcal{H}_{\mathcal{V}}^{T2H}$ & Tail-to-head vertex representations\\
        $\mathcal{H}_{\mathcal{V}}^{SfH}$ & Starting-from-head vertex representations\\
        $t\in T$ & Timestamps\\
        $q$ & The number of vertex groups \\
        $dim$ & Representation dimensions\\
        $\oplus$ & Stack operator\\
        $\varpi$ & Head/tail ratio threshold\\
        $d(\cdot)$ & Short-term trend encoder\\
        $\hat{d}(\cdot)$ & Long-term trend encoder\\
        $deg(\cdot)$ & Degree of vertex\\
        $y\in Y$ & Labels of biased structural evolution \\
        $(v,v^+)$ & Positive vertex pair from same structural evolution group \\
        $(v,v^-)$ & Negative vertex pair from different structural evolution groups\\
        $\mathcal{L}_{class}$ & Classification loss\\
        $\mathcal{L}_{contrast}$ & Contrastive loss\\
        $\mathcal{L}_{DS}$ & Downstream task loss\\
        $\mathcal{L}_{fair}$ & Fairness loss\\
        $\gamma_1, \gamma_2, \gamma_3, \gamma_4$ & Loss hyperparameters\\
        $||\Theta||_1, ||\Theta||_2$ & $\ell_1$, $\ell_2$ norm of model parameters\\ 
        $p_e$ & Probability of edge $e$ existing\\
        $k$ & Top $k$ prediction list\\
    \hline
    \end{tabular}}\vspace{-1em}
\end{table}

\section{Annotating Algorithm for FaT, T2H, and SfH}
\label{sec:annotating_algorithm}
We devise a slope-based algorithm in Algorithm \ref{algorithm:label_strategy}, identifying the degree evolving trend to annotate FaT, T2H, and SfH. $\rho$ is set to 0 in our experiments.

\begin{algorithm}[h] 
\renewcommand{\algorithmicrequire}{\textbf{Input:}}
\renewcommand{\algorithmicensure}{\textbf{Output:}}
	\caption{\small{Slope-based Annotation Algorithm}} 
	\label{algorithm:label_strategy}
	\begin{algorithmic}
	\small
		\Require Vertex set $\mathcal{V}$, timestamp range $T$, head/tail ratio threshold $\varpi$, degree variation threshold $\rho$.
 
		\Ensure Label set $Y$ for $\mathcal{V}$, where $y\in \{FaT, T2H, SfH\}$ is the label of structural evolution for vertex $v\in V$.
        \State
            \State head\_num = $|\mathcal{V}| \times \varpi$
            \State head\_group = sort(deg($\mathcal{V}$))[:head\_num]
            \For {vertex $v$ in $\mathcal{V}$} 
            \If{$v_t$ in head\_group}  
                \State $y_v = SfH$; 
            \Else 
                \State $argmax(deg(v_T))$ //Find the time when $v$ has the biggest degree;
                \State $argmin(deg(v_T))$ //Find the time when $v$ has the smallest degree;
            \If {$argmax(deg(v_T)) - argmin(deg(v_T)) > \rho $}
                \State $y_v = T2H$; 
            \Else
                \State $y_v = FaT$; 
            \EndIf
            \EndIf
            \EndFor
        \State return $Y$
	\end{algorithmic} 
\end{algorithm}


\section{Experiments}
\label{app:sec:complementary_exp}

\subsection{Experiment Configurations and Implementation Details}
\label{sec:exp_settings}
To ease reproductivity, we provide the experiment configurations and implementation details of FairDGE.


For all fair graph learning baselines, including FairGNN, Fair-VGNN, DegFairGNN, and TailGNN, we adopted the source code released by the authors in GitHub and incorporated them into our code framework. The GCN and GAT models are implemented by the \textit{HGCN} \cite{chami2019hyperbolic} framework, and the Euclidean space version is used. For MPNN-LSTM and EvolveGCN, we adopted the implementation in the \textit{PyTorch Geometric Temporal} library. The hidden layer version of EvolveGCN is used in the experiments. The default parameters are employed for all baseline methods, but we tune the learning rate for each baseline to get the best results. 

Our proposed method, FairDGE, is implemented by \textit{PyTorch}. The learning rate is set to 5e-4 after parameter searching. The model dimensions are set the same for FairDGE and all baselines, enabling a fair comparison. The number of GNN layers in FairDGE is set to 2. 
In addition to benchmarking the performance of FairDGE against the original version of the baselines, we added our fair loss $\mathcal{L}_{fair}$ to each baseline and further validated the effectiveness of the fair loss. FairGNN and FairVGNN train a discriminator to eliminate the performance diversity between the advantaged and disadvantaged sensitive group of vertices. We align their methods to train the discriminator for SfH, T2H, and FaT. Since DegFairGNN and TailGNN
We use each user's last interacted item in both datasets as the test data. 

\subsection{Additional Hyperparameter Study}
\label{app:sec:Hyperparameter_study}

\begin{table}[t]
\caption{Different loss hyperparameters.}\label{app:tab:gamma_hyperparameter}\vspace{-1em}
\resizebox{0.45\textwidth}{!}{%
\begin{tabular}{|l|c|ccc|cc|}
\hline
\multicolumn{2}{|c|}{\multirow{2}{*}{}} & \multicolumn{3}{c|}{Performance ($\uparrow$)}                         & \multicolumn{2}{c|}{Fairness ($\downarrow$)} \\ \cline{3-7} 
                               \multicolumn{2}{|c|}{}                                            & \multicolumn{1}{c|}{HR@20}  & \multicolumn{1}{c|}{NDCG@20} & PREC@20 & \multicolumn{1}{c|}{rHR}        & rND       \\ \hline
\multirow{3}{*}{$\gamma_1(\mathcal{L}_{DS})$}       & 0.25                   & \multicolumn{1}{c|}{0.2025} & \multicolumn{1}{c|}{0.0903}  & 0.0101  & \multicolumn{1}{c|}{0.1489}     & 0.2075    \\ \cline{2-7} 
                                                    & 0.75                   & \multicolumn{1}{c|}{0.2037} & \multicolumn{1}{c|}{0.0911}  & 0.0102  & \multicolumn{1}{c|}{0.1545}     & 0.2548    \\ \cline{2-7} 
                                                    & 2.5                    & \multicolumn{1}{c|}{0.2188} & \multicolumn{1}{c|}{0.1009}  & 0.0109  & \multicolumn{1}{c|}{0.1548}     & 0.2610    \\ \hline
\multirow{3}{*}{$\gamma_2(\mathcal{L}_{Class})$}    & 0.25                   & \multicolumn{1}{c|}{0.2025} & \multicolumn{1}{c|}{0.0903}  & 0.0101  & \multicolumn{1}{c|}{0.1489}     & 0.2075    \\ \cline{2-7} 
                                                    & 0.75                   & \multicolumn{1}{c|}{0.2077} & \multicolumn{1}{c|}{0.0928}  & 0.0104  & \multicolumn{1}{c|}{0.1548}     & 0.2627    \\ \cline{2-7} 
                                                    & 2.5                    & \multicolumn{1}{c|}{0.2203} & \multicolumn{1}{c|}{0.1013}  & 0.0110  & \multicolumn{1}{c|}{0.1537}     & 0.2472    \\ \hline
\multirow{3}{*}{$\gamma_3(\mathcal{L}_{contrast})$} & 0.25                   & \multicolumn{1}{c|}{0.2025} & \multicolumn{1}{c|}{0.0903}  & 0.0101  & \multicolumn{1}{c|}{0.1489}     & 0.2075    \\ \cline{2-7} 
                                                    & 0.75                   & \multicolumn{1}{c|}{0.2052} & \multicolumn{1}{c|}{0.0915}  & 0.0103  & \multicolumn{1}{c|}{0.1541}     & 0.2551    \\ \cline{2-7} 
                                                    & 2.5                    & \multicolumn{1}{c|}{0.2181} & \multicolumn{1}{c|}{0.1038}  & 0.0109  & \multicolumn{1}{c|}{0.1561}     & 0.2711    \\ \hline
\multirow{3}{*}{$\gamma_4(\mathcal{L}_{fair})$}     & 0.25                   & \multicolumn{1}{c|}{0.2025} & \multicolumn{1}{c|}{0.0903}  & 0.0101  & \multicolumn{1}{c|}{0.1489}     & 0.2075    \\ \cline{2-7} 
                                                    & 0.75                   & \multicolumn{1}{c|}{0.2198} & \multicolumn{1}{c|}{0.1019}  & 0.0110  & \multicolumn{1}{c|}{0.1543}     & 0.2551    \\ \cline{2-7} 
                                                    & 2.5                    & \multicolumn{1}{c|}{0.2205} & \multicolumn{1}{c|}{0.1004}  & 0.0110  & \multicolumn{1}{c|}{0.1535}     & 0.2447    \\ \hline
\end{tabular}}
\end{table}

Additional hyperparameter testing on loss hyperparameters is presented in this section. The results reported in Table \ref{app:tab:gamma_hyperparameter} test the impact of different loss hyperparameters $\gamma_1, \gamma_2, \gamma_3, \gamma_4$ in Eq. (\ref{eq:final_loss}) on the performance and fairness scores on the Amazon Book dataset. Specifically, we respectively tune $\gamma_1$, $\gamma_2$, $\gamma_3$, and $\gamma_4$ from $\{0.25, 0.75, 2.5\}$. At each round, we only tune one and fix the remaining three to $0.25$. For example, when tuning $\gamma_1$ from $\{0.25, 0.75, 2.5\}$, we fix $\gamma_2=\gamma_3=\gamma_4=0.25$. Results in Table \ref{app:tab:gamma_hyperparameter} demonstrate that FairDGE is not sensitive to $\gamma$ on both performance and fairness. When all the $\gamma$ are 0.25, the results are the best on the fairness score.

\section{Complexity Analysis}
FairDGE consists of two modules, i.e., trend-aware structural evolution learning and dual debiasing for encoding structurally fair embeddings. We analyze the computational complexity of each module, respectively.

Trend-aware structural evolution learning consists of long-short-term degree trend encoding, structural evolution embedding, and a bias classification task. The computational complexity is $O(max(L \cdot |\mathcal{V}|^2,|T|\cdot dim_{hid}^2+|T|\cdot dim_{hid}\cdot dim))$ according to \cite{blakely2021time, rotman2021shuffling}. 
$|\mathcal{V}|$ is the number of vertices, $L$ is the number of GNN layers, $|T|$ is the number of snapshots of the dynamic graph $\mathcal{G}$, $dim_{hid}$ is the hidden dimension of GRU and $dim$ is the representation dimension of our model. Since $|T| \ll |\mathcal{V}|$ and $L \ll |\mathcal{V}|$, the overall complexity of trend-aware structural evolution learning is $O(|\mathcal{V}|^2)$, approximately. 

In the dual debiasing module, contrastive learning with negative sampling is first employed. Its computational complexity is $O(|\mathcal{V}|^2)$. The complexity of the followed fairness loss and the link prediction task are both $O(|\mathcal{E}|)$. Therefore, the complexity of the dual debiasing module is $O(max(|\mathcal{V}|^2,|\mathcal{E}|))$.

Since the FairDGE trained in end-to-end manner, the overall computational complexity is $O(max(|\mathcal{V}|^2,|\mathcal{E}|))$.

\end{document}